\pdfoutput=1
\documentclass[11pt]{article}
\usepackage{hyperref}
\usepackage{url}
\usepackage{amsmath}        % 数学公式基础包
\usepackage{amsfonts}       % 字体符号包
\usepackage{amssymb}        % 数学符号包
\usepackage{mathrsfs}       % 花体字母包
\usepackage{bm}             % 数学粗体包
\usepackage{amsthm}         % 数学定理包
\usepackage{graphicx}
\usepackage{multirow}
\usepackage{subcaption}
\usepackage{graphicx}
\usepackage{microtype}
\usepackage{multirow}
\usepackage{footmisc}
\usepackage{soul}
\usepackage{arydshln}

\usepackage{algpseudocode}
\usepackage{xspace}
\usepackage{times}
\usepackage{latexsym}
\usepackage{graphics}
\usepackage{graphicx}
\usepackage{color}
\usepackage{colortbl}
\usepackage{amssymb}
\usepackage{amsmath}
\usepackage{bbm}
\usepackage{booktabs}
\usepackage{arydshln} %负责画虚线的包
\usepackage[most]{tcolorbox}
\usepackage{hyperref}

\usepackage{xcolor}
\definecolor{myorange}{RGB}{2, 142, 2}
\usepackage[linesnumbered,ruled,vlined]{algorithm2e}
\usepackage{latex/acl}

\usepackage{times}
\usepackage{latexsym}

\usepackage[T1]{fontenc}
\usepackage[utf8]{inputenc}

\usepackage{microtype}

\usepackage{inconsolata}

\title{How Abilities in Large Language Models are Affected by Supervised Fine-tuning Data Composition}

\author{Guanting Dong\thanks{\ \ Work done during internships at Alibaba Group.},  Hongyi Yuan\textsuperscript{*}, Keming Lu, Chengpeng Li\textsuperscript{*}, Mingfeng Xue \\ 
\textbf{Dayiheng Liu, Wei Wang, Zheng Yuan, Chang Zhou, Jingren Zhou}\\
Alibaba Group \\
\texttt{\{dongguanting.dgt,yuanzheng.yuanzhen,ericzhou.zc\}@alibaba-inc.com} \\
}

\begin{document}
\maketitle
\begin{abstract}

Large language models (LLMs) with enormous pre-training tokens and parameters emerge diverse abilities, including math reasoning, code generation, and instruction following.
These abilities are further enhanced by supervised fine-tuning (SFT).
While the open-source community has explored ad-hoc SFT for enhancing individual capabilities, proprietary LLMs exhibit versatility across various skills. Therefore, understanding the facilitation of multiple abilities via SFT is paramount. 
% In this study,
% The instruction-following ability of large language models (LLMs) heavily relies on high-quality, diverse, and complex training samples during the supervised fine-tuning (SFT) stage for effective generalization.
% However, the lack of quantitative analysis on the data composition problem from different sources hinders us from further enhancing the capabilities of LLMs in a comprehensive manner.
In this study, we specificially focuses on the interplay of data composition between mathematical reasoning, code generation, and general human-aligning abilities during SFT.
% and further pose four different research questions.
We propose four intriguing research questions to explore the association between model performance and various factors including data amount, composition ratio, model size and SFT strategies. Our experiments reveal that distinct capabilities scale differently and larger models generally show superior performance with same amount of data.
Mathematical reasoning and code generation consistently improve with increasing data amount, whereas general abilities plateau after roughly a thousand samples. 
% Moreover, the directly mixed sources in the SFT phrase exhibit improved performance in low-resource scenarios and a decline in high-resource scenarios when compared to individual domains. 
Moreover, we observe data composition appears to enhance various abilities under limited data conditions, yet can lead to performance conflicts when data is plentiful.
% We find mixing data sources from various abilities and applying multi-task learning leads to ability improvements with low data amounts and ability conflicts with high data amounts compared to training each ability individually.
%--- Notably, the phenomenon of low data amounts gain becomes more prominent in general and math abilities as the model parameter size increases.
% We also gain valuable insights into the key factors contributing to the data composition problem.
Our findings also suggest the amount of composition data influences performance more than the composition ratio.
In analysis of SFT strategies, we find that sequentially learning multiple skills risks catastrophic forgetting. 
Our proposed \textbf{Dual-stage Mixed Fine-tuning (DMT) strategy} offers a promising solution to learn multiple abilities with different scaling patterns.
% Experimental results demonstrate that our DMT strategy effectively alleviates performance conflicts and catastrophic forgetting during the SFT phase
% , offering a promising solution to learn multiple abilities with different scaling patterns.

\end{abstract}

\section{Introduction}

% Human aligned large language models need various abilities including math reasoning, coding, general ...

% SFT has its sc law.
% Different abililty have different sc law.
% Different ability could conflict each other. Data ratio could be important.

% We investigate the performance among different data ratio / data size / model size.

% Pretrained large language models have learned knowledge, 
% Supervised fine-tuning 

% Large language models (LLMs) have demonstrated remarkable proficiency in tasks that pertain to reasoning, coding, commonsense, and world knowledge \cite{}. 
Recent research has demonstrated the remarkable and versatile proficiency of large language models (LLMs) in dealing with a variety of real-world tasks expressed in natural languages \citep{instructgpt,palm2,openai2023gpt4,luo2023chatkbqa}, especially Information Extraction (IE) \cite{lu2022unified,xu2023large,zhao2023demosg,cheng2023accelerating,wang2023gptner,zhang2024linkner,li2024matching}, Information Retrieval (IR) \cite{zhu2024large,10.1145/3589335.3641299} and Spoken Language Understanding (SLU) \cite{hoscilowicz2024large,yin2024large,cheng2023ml,cheng2024towards,dong2023demonsf}.
Among the tasks, LLMs especially emerge with three outstanding abilities in reasoning \citep{gsm8k,cot}, coding \citep{chen2021codex}, and aligning general human intentions \citep{instructgpt}, which have drawn much attention from the LLM research community. 
In order to further incentivize such abilities, it necessitates supervised fine-tuning (SFT) stages on annotated task data. 

However, existing research has mostly conducted separate SFT investigations on each of the three tasks, where reasoning and coding abilities require SFT on in-domain human-annotated or augmented data \citep{rft,wizardcoder,yu2024metamath} while diverse and complex human instructions are applauded for aligning human intentions \citep{wang2023selfinstruct,alpaca,cheng2023mrrl,xu2023wizardlm,lima,openchat,instag}. 
As shown by the strong performance of proprietary LLMs such as GPT-4 \citep{openai2023gpt4} and Claude, LLMs have the potential to master all the tasks in one model. 
Therefore, it is of paramount importance to investigate the versatile performance of SFT with composite task data, and understanding and addressing the challenges posed by the data composition problem in the SFT stage is crucial for further enhancing the capabilities of LLMs in a comprehensive manner.

% The instruction-following ability of LLMs is obtained from the supervised fine-tuning (SFT) stage which heavily relies on high-quality, diverse, and complex training samples for effective generalization \cite{}. 
% The training data for the SFT stage can consist of a mixture of data from multiple sources. 
% These different sources of SFT data have different qualities, quantities, and topics which raises the problem of how to compose these SFT data.
% Relying on heuristics for filtering and empirical mixing strategies cannot fundamentally address the conflicts between different ability items \cite{}.
% Therefore, understanding and addressing the challenges posed by the data composition problem in the SFT stage is crucial for further enhancing the capabilities of LLMs in a comprehensive manner.

% Different abilities may need different amounts of SFT data to be activated. 
% We want to understand how the amount of data and the data composition influence each ability.
% We use math, code, and general instructional following as the 研究对象
% Math and coding show strong data requirements during SFT, while general ability may be less is more.
% Combining SFT data from multiple sources/abilities may contradict or enhance each other.
% In this paper, we empirically investigate different LLM abilities under different SFT data compositions... 

In essence, the tasks of reasoning, coding, and aligning human intentions are of different characteristics. Reasoning and coding tasks require ad-hoc abilities of complex and detailed logic in decomposing task instructions and dealing with non-linguistic and symbolic features \citep{chen2021codex,huang-chang-2023-towards}, whereas aligning human intentions requires versatility and understanding obscure intentions expressed in human instructions \citep{instag}. 
Given the fundamental difference among the tasks, multi-task learning with composite data fine-tuning for small-scaled pre-trained language models is prone to catastrophic forgetting \citep{9349197}, hindering the fine-tuned performance of one model on separate tasks. 
Many efforts have been made to compensate for the phenomenon \citep{liang2021rdrop,xu-etal-2021-raise,yuan-etal-2023-hype}. 
There has also been research discovering that scaling up the pre-trained language model scale and the fine-tuning data scale are beneficial for zero-shot out-of-domain generalization on various linguistic tasks while leaving out the assessment of in-domain performance \citep{sanh2022multitask,flan,longpre2023flan}. 
Given the increased capacity of LLMs, the multi-task performance by SFT on composite data of essentially different downstream tasks is less studied. 
Understanding the SFT performance with composite data and corresponding scaling patterns is of great utility in practice. 

% Recently, a series of studies on SFT have demonstrated that LLMs have achieved a well-balanced set of fundamental understanding abilities, including language comprehension, semantic understanding, and syntactic parsing \cite{}, which enable the models to perform exceptionally well across a wide range of NLP benchmarks. 
% However, when delving deeper into the SFT stage, specific reasoning skills, such as mathematical reasoning and code generation, pose unique challenges due to their distinct data characteristics and high logical complexity.
% Several studies \cite{} have highlighted the limitations of extending these capabilities, showing that injecting non-linguistic skills often comes at the cost of losing general linguistic skills.
% Akin to all neural models, language models are susceptible to catastrophic forgetting\cite{} when trained for multi-task learning, especially when the two tasks are substantially different - linguistic vs. non-linguistic. Although certain studies \cite{} have examined the data composition problem of different skills in the pre-training phase from the perspective of scaling laws, the lack of understanding of the SFT stage hinders our progress toward practical applications.

In this study, we focus on the data composition problem among \textbf{mathematical reasoning}, \textbf{code generation}, and \textbf{general human-aligning abilities} in SFT. 
We aim to comprehensively investigate the relationship between model performance and different factors including data amount, data composition ratio, model scales, and SFT training strategies. 
We also investigate how the relationship varies under different scales. 
Specifically, we focus on the following four research questions:

\textbf{1.}~ \textit{How do math reasoning, coding, and general abilities scale with SFT data amounts?}

\textbf{2.}~ \textit{Are there performance conflicts when combining these three abilities in SFT?}

\textbf{3.}~ \textit{What are the key factors that induce the performance conflicts?}

\textbf{4.}~ \textit{What are the impacts of different SFT strategies for composite data?}
%---------mark

%-----------mtbench all------------
\begin{figure*}[t]
    \centering
    \small    \includegraphics[width=\linewidth]{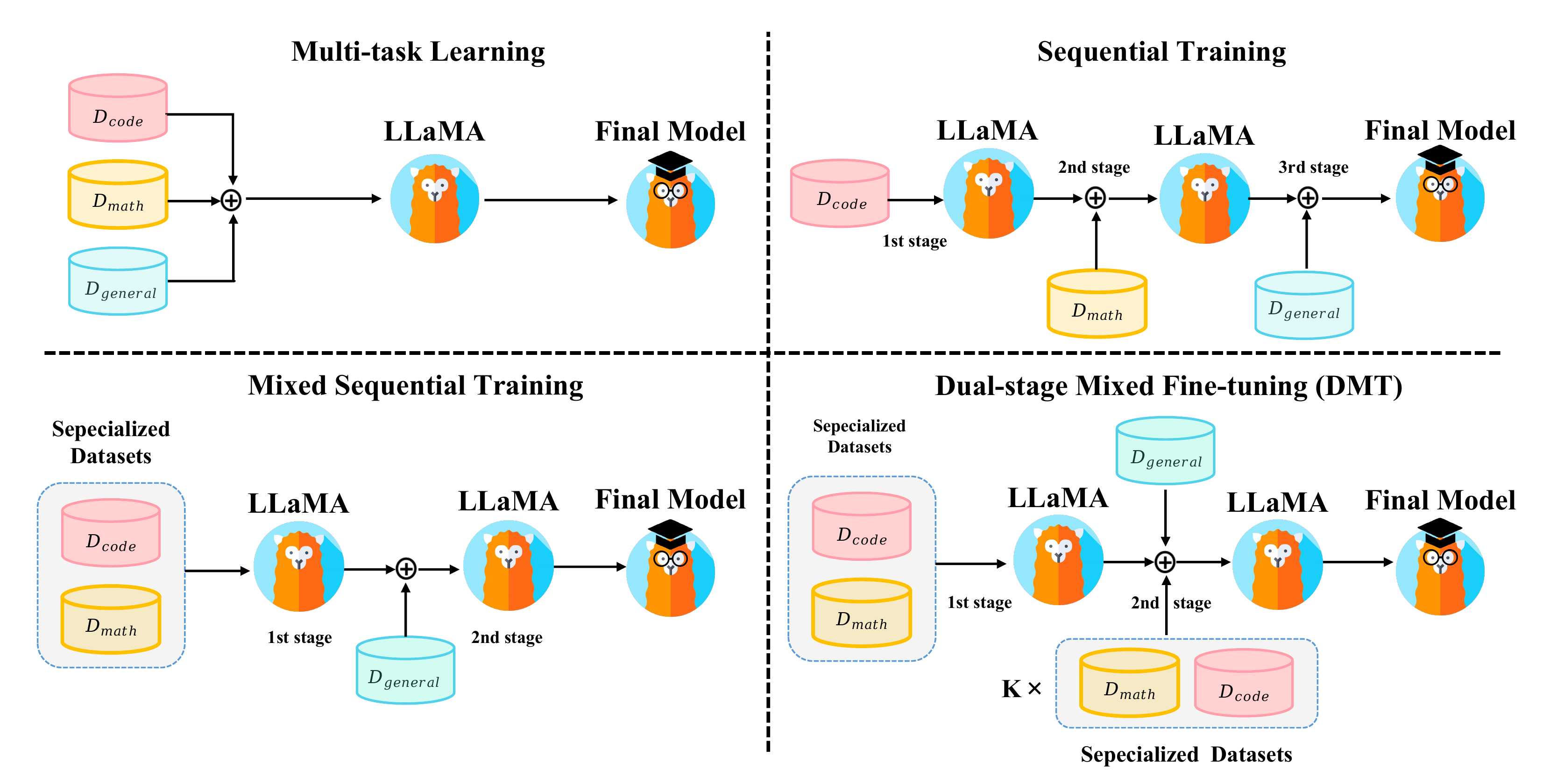}
      \vspace{-0.4cm}
    \caption{The illustration of four different training strategies in this paper.}
    \label{fig:4_train_strategies}
    % \vspace{-0.2cm}
\end{figure*}
%-----------mtbench alls------------
To answer these questions, we conduct experiments on three benchmarks, which are GSM8K \citep{gsm8k} for mathematical reasoning, HumanEval \citep{chen2021codex} for coding, and MT-Bench \citep{zheng2023judging} for general human alignment. 
We fine-tune LLMs on the related training data to activate these abilities.
% except for MT-Bench using ShareGPT.
Furthermore, we conduct extensive analysis regarding model parameter scales ranging from LLaMA 7B to 33B \citep{touvron2023LLaMA} and explore four different SFT strategies shown in Figure~\ref{fig:4_train_strategies}: multi-task learning, sequential training, mixed sequential training, and dual-stage mixing fine-tuning (DMT), providing empirical guidance for learning a versatile LLM with composite SFT.
The key findings of this paper can be summarized as follows:

\begin{itemize}

\item Different SFT abilities exhibit distinct scaling patterns, while larger models show better performances with the same data amount generally.

\item Compared to single ability learning, multi-task learning multiple abilities exhibits improvement in low-resource and decline in high-resource. 
Additionally, as the model size increases, there is a greater performance gain in low-resource settings for math and general abilities.

\item Data amounts directly influence each ability, while the data ratio is insignificant.

%多任务训练会导致不同SFT能力间产生性能冲突,而两种顺序训练方法则会产生对sepecialized abilities的灾难性遗忘,我们提出的DMT策略可以同时有效地缓解性能冲突与灾难性遗忘,并通过调节k实现通用能力与特定能力性能的平衡,满足不同的训练需求
\item Multi-task learning lead to conflicts, while sequential training results in catastrophic forgetting. Our proposed DMT effectively alleviates both performance conflicts and catastrophic forgetting in the SFT phrase, achieving a balance between general and specialized abilities.
% \item Sequential training and mixed-first then general strategies better preserve general abilities compared to direct mixing. However, they may lead to some knowledge forgetting in coding and mathematical reasoning abilities. Conversely, an advanced strategy of mixed continue training can retain all three abilities as much as possible, providing empirical guidance to alleviate the data composition problem in the SFT phase.

% substantially better arithmetic while retaining their linguistic prowess. 尽可能保留语言能力,更好的算数

\end{itemize}

\section{Related Works}

\paragraph{Supervised Fine-Tuning of Large Language Models}
Large Language Models (LLMs) have shown notable zero-shot performance in various domains \cite{brown2020language,wu2021yuan,hou2024large,dong2023revisit,zhou2024grasping,wu2023semantic,song2023large}, prompting further development to push the boundaries of these models. To delve deeper to their potential, LLMs are subjected to a Supervised Fine-Tuning (SFT) phase, enhancing their ability to solve tasks and align better with human instructions. Here, we extend the conventional definition of SFT to include various forms of sequence-to-sequence fine-tuning, such as fine-tuning for human alignment, instruction following, and domain-specific task optimization \citep{zhou2023instructionfollowing,yuan2023rrhf,cheng2023m,zhang2024instruction}.

Recent research has delved into multi-task instruction fine-tuning of pre-trained LLMs to bolster their zero-shot performance across numerous downstream NLP tasks \citep{sanh2022multitask}. In an effort to encompass existing NLP tasks comprehensively, \citeauthor{flan,longpre2023flan} curated the expansive FLAN dataset specifically for instruction-based fine-tuning. LLMs, both open-source \citep{chung2022scaling} and proprietary \citep{singhal2022large}, fine-tuned with FLAN, have demonstrated enhanced zero-shot performance on a variety of unseen tasks.

While research has probed into the generalization capabilities of LLMs within out-of-distribution domains \cite{liu2024good,yuan2024revisiting,wang2024multiperspective}, the effects of multi-task training on in-domain performance remain under-explored. With the ascent of proprietary models like ChatGPT, the focus on SFT for aligning LLMs with human intent has intensified \citep{ouyang2022training}. Moving away from crowd-sourced SFT data, recent initiatives have generated SFT datasets from user logs within proprietary LLM platforms \citep{vicuna2023,openchat}, employing the models themselves to assist in the data generation process \citep{wang2023selfinstruct,alpaca,cheng2023accelerating,lei2023instructerc,xu2023wizardlm,xue2023occuquest}. Additionally, methods to improve SFT data quality have been proposed, targeting more accurate alignment with human interactions \citep{lima,tulu,instag,liu2024makes}.

Furthermore, SFT has proven beneficial for LLMs in specialized areas such as mathematical reasoning \citep{gsm8k,hendrycks2021measuring,rft,chen2024autoprm,mammoth,gou2024tora,li2023query,yue2024mammoth2} and code generation tasks \citep{codealpaca,wizardcoder,wang2024dolphcoder,wei2023magicoder}. Taking advantage of their advanced interactive capabilities, some researchers have leveraged supervised fine-tuned LLMs to compose commands that interface with external tools, thus enhancing the handling of assorted downstream applications \cite{shen2023hugginggpt,yao2023react,yao2023tree,song2024knowledge,fu2024preact}. This paper examines the SFT performance using composite datasets, considering different model sizes and data amounts.

\paragraph{Scaling Laws in Large Language Models} 

The exceptional performance of LLMs comes from scaling up model sizes, data amounts, and computational costs to massive scales. 
Therefore, it is crucial to explore the model performance across an exponential range of scales. 
Many endeavors have been made to discuss the scaling laws for pre-training \citep{palm2,hoffmann2022training}, transfer learning \citep{chronopoulou2019embarrassingly}, preference modeling \citep{gao2022rmscaling} and mathematical reasoning \citep{rft}. In this paper, we also explore the SFT performance with composite data from the perspective of different scales of model sizes and data amounts.

% Large language models obtain the ability to follow human instructions through supervised fine-tuning (SFT) \citep{instructgpt,longpre2023flan,alpaca,vicuna2023,instag,lima,openchat}.
% SFT data could come from multiple resources to activate multiple abilities including but not limited to code \citep{codealpaca,wizardcoder}, math \citep{gsm8k,hendrycks2021measuring,rft,mammoth}, and other general human instruction following ability \citep{vicuna2023,2023openassistant,xu2023wizardlm,ding2023enhancing}.
% \cite{longpre2023flan,tulu,instag} perform SFT on a mix of instruction datasets and perform well among multiple benchmarks.
% This paper is focused on investigating how SFT data composition influences the different abilities in large language models.

% Transfer learning and life-long learning
% \citep{longpre2023flan,t0}

\section{Experiments}
We have SFT datasets $\{D_1,D_2,...,D_k\}$ where each $D_i=\{q_{i,j},r_{i,j}\}_j$ contains queries and responses from one source.
We consider each SFT dataset to correspond to one ability and we also have $k$ in-domain metrics to measure them.
We investigate the performances of in-domain metrics with different dataset compositions ($D\subset \cup_{1\leq i \leq k}D_i$) and training strategies on different sizes of LLMs.
% The goal of this paper is to try to understand the scaling relationship between model performance and different factors, including \textbf{data amount}, \textbf{data ratio}, \textbf{model parameters}, and \textbf{data composition strategies} in SFT phrase. We expect that the LLM $\rho$ of the pre-trained model can be Learning mathematics, coding, and general human-aligning abilities in SFT dataset $\mathcal{D}$. The dataset is defined by $\mathcal{D}=\{q_i,a_i\}_i$, where $q$ is a question which includes task description and query, $a$ is an answer.
% We perform supervised fine-tuning on dataset $\mathcal{D}$ to obtain an SFT model $\pi$.
% We use $\pi$ to generate reasoning paths and answers in the test set by greedy decoding and report the accuracy (i.e. $acc$ or $F1$) as our metric here.

% This section introduces the experimental settings,
% results and analysis. We answer the following research
% questions (RQs). \textbf{RQ1:} What are the individual scaling curves for coding, mathematical reasoning, and General abilities? \textbf{RQ2:} Is there a performance conflict when combining these three abilities? \textbf{RQ3:} What is the key factor cause performance conflicts among capability items, data volume or ratio? \textbf{RQ4:} What is the impact of different data composition strategies for performance conflicts?

\subsection{Experiment Setup}
We collect three SFT datasets $\{D_1,D_2,D_3\}$ including GSM8K RFT \citep{rft}, Code Alpaca \citep{codealpaca}, and ShareGPT \citep{vicuna2023} to represent math reasoning, coding, and general human-aligning ability SFT dataset respectively. 
We will integrate a new SFT dataset $D$ by these three datasets to investigate how data composition affects the model performances.
We use GSM8K test set \citep{gsm8k}, HumanEval \citep{chen2021codex}, and MT-Bench \citep{zheng2023judging} to measure abilities including math reasoning, coding, and general human-aligning. 
We use LLaMA \citep{touvron2023LLaMA} series as our pretrained language models and use FastChat framework \citep{zheng2023judging} for fine-tuning. 
We fine-tune models with 3 epochs and a peak of 2e-5 learning rate. The batch size during SFT is 16. More details about SFT datasets, evaluation metrics, implementations and Training FLOPs can be found in Appendix \ref{sec:app_a}, \ref{sec:app_b}, \ref{sec:app_c} and \ref{sec:app_d}.

% \subsection{RQ1: What are the individual scaling curves for Code, Math, and general abilities?}
\subsection{RQ1. Individual Ability Performance vs. Data Amount}

% In this section, we gain insight into the scaling relationship between the individual model performance and the data amount of the three abilities.
% According to FLAN \citep{flan}, increasing the number of tasks in the SFT phase results in improved overall performance. 
The instruction following ability can be activated via SFT on datasets like ShareGPT which contain around 100 thousand samples.
However, \cite{lima} demonstrates that strong base models can achieve human alignment with just 1000 samples. 
% It is important to examine how the amount of SFT data relates to human alignment performance. 
Specialized abilities such as math reasoning require a large amount of data \citep{gsm8k,rft}, unlike general abilities. Therefore, it is crucial to investigate how each ability improves as the data amount increases.

\textbf{Experimental Design:} We conduct SFT on LLaMA of various sizes using \{1, 1/4, 1/16, 1/64, 1/256\} proportions of the training set obtained from GSM8K RFT, Code Alpaca, and ShareGPT seperately. This allowed us to evaluate each ability with various data sizes and model sizes.
% the models' performance in the areas of mathematics, coding, and general human-aligning ability. 
% Additionally, we further investigate the impact of increasing the parameter size from 7B to 33B on the scaling relationship.
%-----------single scaling------------
\begin{figure}[t]
    \centering
    \small    \includegraphics[width=\linewidth]{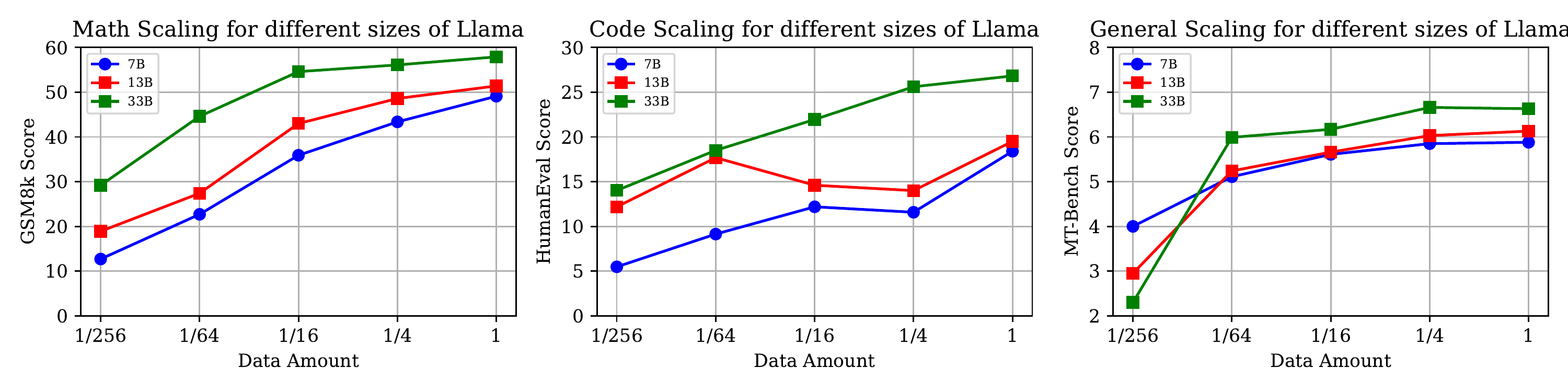}
    \caption{The scaling curve of different sizes of LLaMA in three individual domains.}
    \label{fig:single_scaling}
\end{figure}
%-----------single scaling------------

\textbf{Results and Analysis.} 
% Figure1 呈现了不同能力项在SFT阶段的data scaling curve.我们有以下发现
% (1)不同的能力项有不同的scaling Laws
% 具体来讲,数学能力在各个模型size下均与数据量呈现正相关的趋势,随数据量增加而提升.
% 通用instruction-following能力同样呈现近乎单增的scaling curve,不同的是,通用能力仅仅需要1k左右的数据就产生涌现的能力(1/256至1/64),而当到达一定阈值之后(1/64),性能逐渐趋近于平稳,这也进一步佐证了LIMA less is more的结论,少量的高质量语料是大模型获得通用能力涌现的关键.除此以外,Code能力在模型参数量较小时scaling曲线呈现irrational的情况,然而当其参数量增加到33B时,性能与数据量则呈现近似log linear的scaling关系,这可能暗示当模型参数量提升至一定阈值时,模型对代码的理解和生成能力具有一定的鲁棒性,从而能够更好地捕捉代码的结构和语义.
% (2)当数据饱和的情况下(100%),模型的性能在三种不同的能力上均稳定提升
% (3)在ShareGPT的极低资源下(1/256),模型参数量型越大,其通用能力越弱,可能的原因弱监督信号让大量的参数进行充分学习,这也意味着在特定情况下适当减少模型参数量可能更有利于提高通用能力.
Figure \ref{fig:single_scaling} shows the individual data scaling curves for different abilities after SFT. 
We find that: 
\textbf{Different abilities exhibit different scaling curves.} 
To be more specific, mathematical reasoning capability shows a positive correlation with the data amount across various model sizes which is consistent with \cite{rft}. 
Similarly, general human-aligning ability demonstrates an almost monotonically increasing scaling curve. 
However, it is noteworthy that general ability emerges with only around 1k data samples (ranging from 1/256 to 1/64), and after reaching a certain threshold (1/64), their performances improve slowly. This further supports \cite{lima}, indicating that a small amount of high-quality SFT data is possible for the emergence of general human-aligning ability in LLMs. 
On the other hand, code ability exhibits an irregular scaling curve when the model's parameter count is small (7B \& 13B). However, when the parameter count increases to 33B, its coding performance shows an approximately log-linear trend with the data amount. 
One possible explanation is that Code Alpaca and the samples in HumanEval have different distributions. Larger models can capture shared knowledge across code data distributions in the in-domain samples, which enables them to exhibit some level of generalization to out-of-distribution (OOD) samples.
Another observation is \textbf{larger models show better performances with the same data amount generally.} 
The outlier is with very little data (1/256), smaller models may outperform larger models.
If there is enough data, larger models have stable better performances.
% This finding may implies that, once the model's parameter count exceeds a specific threshold, the model possesses a certain robustness in understanding and generating code, thereby enhancing its ability to capture the intricacies of code structure and semantics.
% When the data is saturated (at 100\% capacity), the model's performance consistently improves in all three ability domains.
% Under extremely low resource settings(ShareGPT 1/256), larger model sizes exhibit weaker general human-aligning ability. This could be attributed to the weak supervision signal, which allows a large number of parameters to undergo sufficient learning. This implies that reducing the model's parameter count appropriately in specific scenarios might be beneficial for enhancing general ability.

% \textbf{Key Findings}

%-----------mtbench all------------
\begin{figure}[h]
    \centering
    \small    \includegraphics[width=\linewidth]{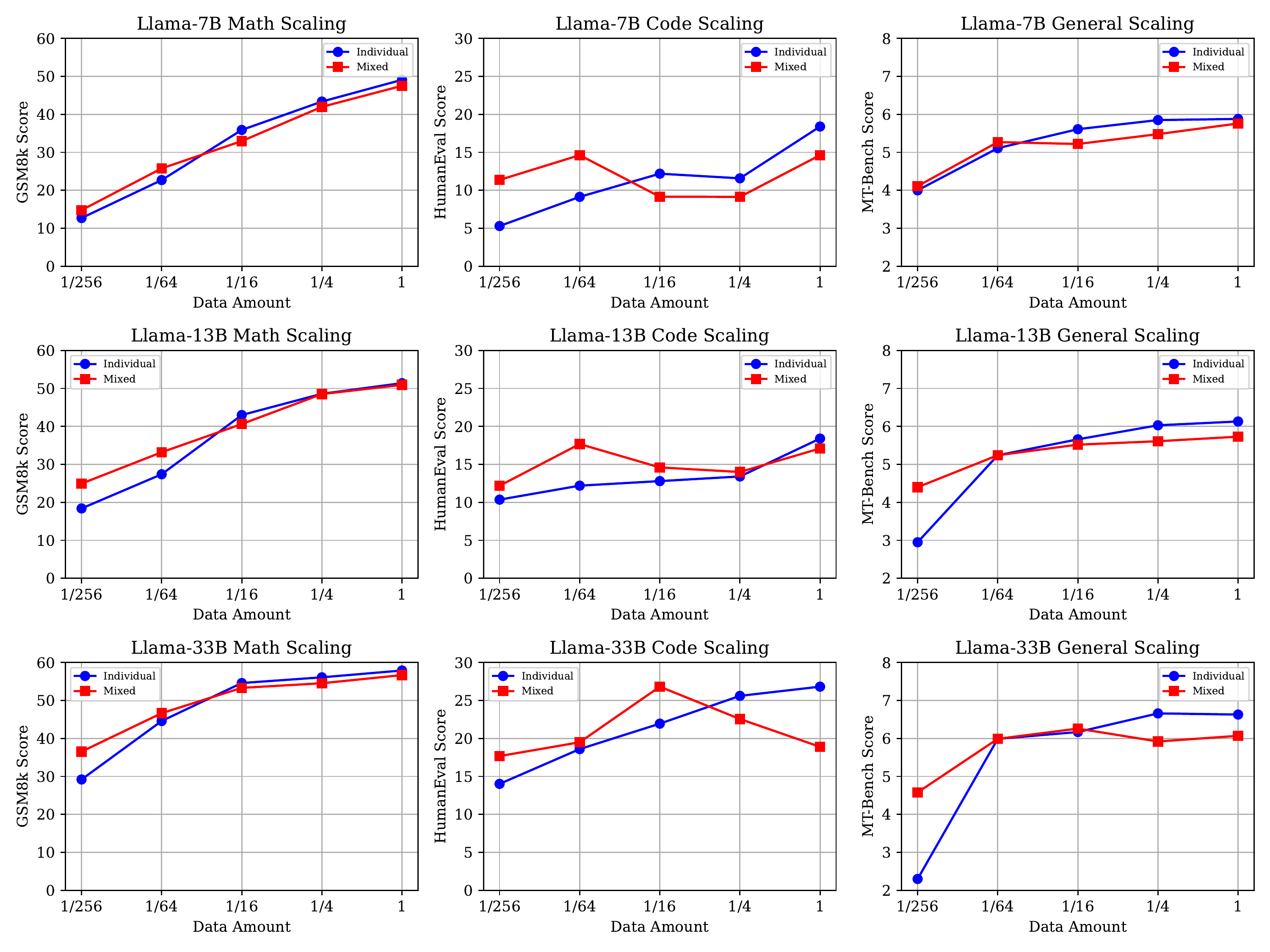}
    \caption{Comparative experiments between mix domains and individual domains for LLaMA.}
    \label{fig:single_mix_scaling}
\end{figure}
%-----------mtbench alls------------

\subsection{RQ2. Performance Difference vs. Mixed Data Amount}

%-----------mix figures------------

%为
% 在7B时,相比于single,mix后模型三种能力项均呈现呈现高资源冲突,低资源增益
% 然而随着模型参数增加,高资源冲突逐渐缓解,低资源的增益逐渐增加

We should deliver a versatile model that requires us to mix various SFT datasets and apply SFT. 
We want to ask how each ability varies due to SFT dataset mixtures.
We investigate it with different amounts of mixed data and compare them with individual ability performance.

% In this section, our objective is to investigate whether there will be performance conflicts after mixing data sources of three kinds of capabilities. Specifically, we conducted the following data scaling comparative experiments on single-source and mixed-source settings.

\textbf{Experimental Design:} For the individual source setting, consistent with the setup in RQ1, we performed fine-tuning on LLaMA models of different sizes using \{1, 1/4, 1/16, 1/64, 1/256\} amounts of training data from GSM8K, Code Alpaca, and ShareGPT separately. 
For the mixed source setting, we sampled \{1, 1/4, 1/16, 1/64, 1/256\} amounts of training data from GSM8K, Code Alpaca, and ShareGPT, and directly mixed them according to the corresponding proportions. In this way, we constructed datasets with fixed proportions of different ability domains, while varying the total data amount. These datasets are then used for fine-tuning the LLaMA models \footnote{We also conduct "Equal Data Amount VS. Equal Data Proportion" experiments in Appendix \ref{sec:app_h}}.

\textbf{Results and Analysis.} Figure \ref{fig:single_mix_scaling} presents results of LLaMA of different sizes on three benchmarks under the individual source and mixed source settings. The following observations are made: \textbf{Abilities are improved with low-resource and are decreased with high-resource compared to individual source abilities.}
In the case of LLaMA-7B, compared to the data scaling curve of the individual source setting, the models fine-tuned with mixed source data consistently demonstrated performance conflicts among the three ability domains at high resources (100\%).
However, as the data volume decreased, a turning point in performance is observed between the two settings in the data range of 1/64 to 1/16. Notably, the models fine-tuned with mixed source data exhibited performance gains at low resources (1/256), indicating that SFT data from different sources benefit each other in a low-resource setting. However, when there is enough data, data from other sources could be viewed as noise for in-domain generalization.
\textbf{As the model size increases, the performance gain in low-resource settings also increases for math and general abilities.} In the case of the 13B and 33B models, it is obvious that the scaling curve for the mix source setting follows a similar trend observed in previous analyses, with the presence of performance intersection points as the data volume scales. However, a crucial distinction arises, whereby larger models exhibit more pronounced performance gains under low resources as the size of model parameters increases. The outlier is the LLaMA-7B (code only, 1/256). 
A possible reason is the introduction of a small amount of unseen code data easily disrupts the original code ability of the pretrained model, as supported by its low HumanEval score (less than 6).
% Consequently, this indicates that this model lacks preliminary code proficiency. 
In conclusion, our finding implies that larger language models excel in acquiring general and specialized abilities from diverse data sources under low-resource conditions \footnote{To validate the generalizability of our conclusions, we further conduct the more experiments on \textbf{World Knowledge},\textbf{ Language Understanding} and \textbf{Translation} in Appendix \ref{sec:app_e}.}.
%-----------RQ3 data ratio------------
\begin{figure}[t]
    \centering
    \small    \includegraphics[width=\linewidth]{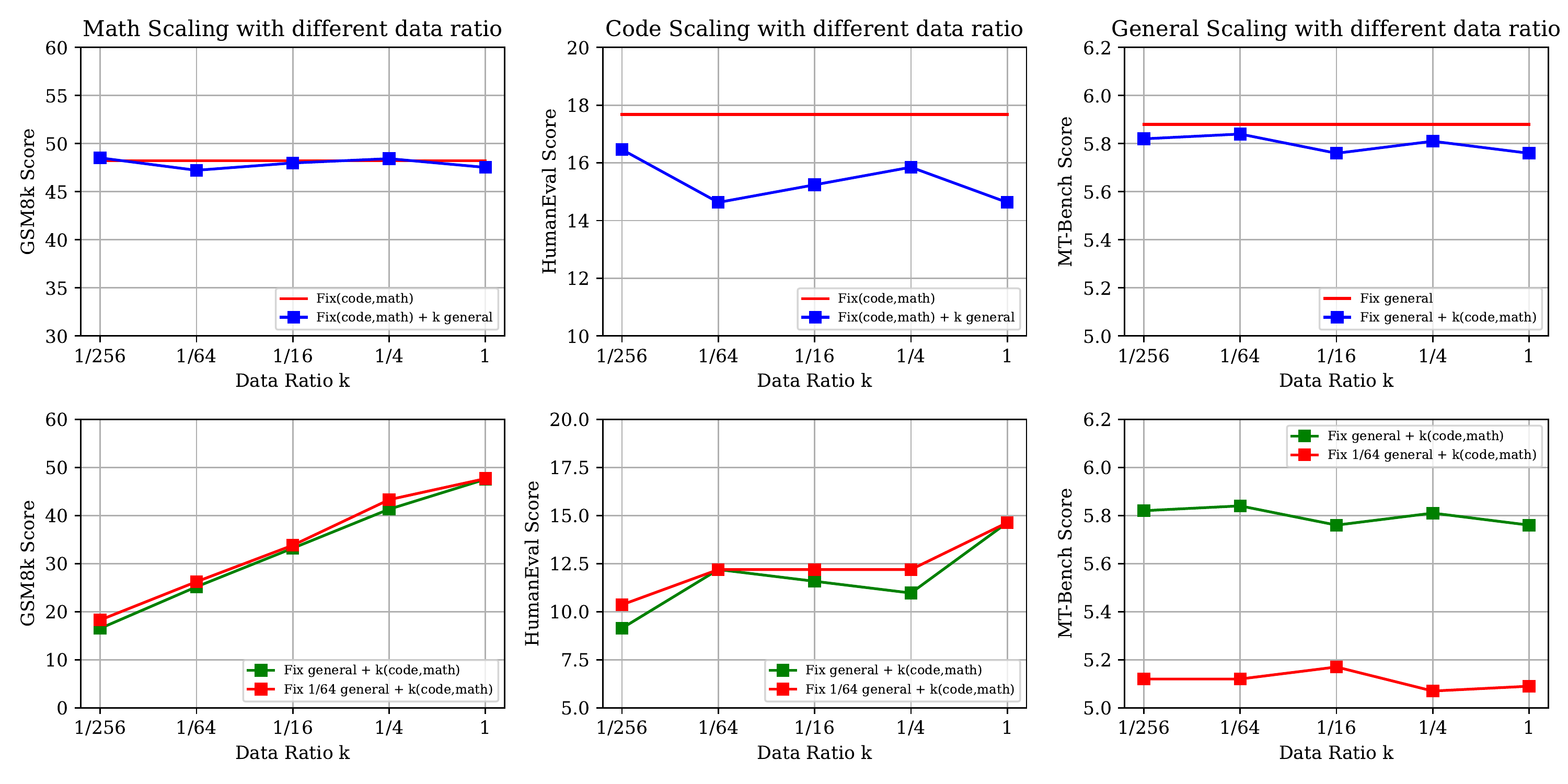}
    \caption{Different data ratio (k) between specific abilities and general abilities on three benchmarks.}
    \label{fig:data ratio}
\end{figure}
%-----------RQ3 data ratio------------

\subsection{RQ3. Performance Difference vs. Data Composition Ratio}
% 固定general scaling code math的设置,在gsm8k和human eval上有明显波动, 在mtbench上波动并不大,这说明code math的数据配比减少,并不会减少xxxx.
%同理固定code math scaling general的设置

We observe ability conflicts in high-resource settings, and we want to investigate the reasons why the conflicts occur.
Two possible factors are the \textbf{data amount} of other abilities is too high or the \textbf{data ratio} of other abilities is too high.
Here we conduct experiments to investigate the data ratio factor.
% To answer this question, we conduct experiments to investigate these two factors.

% We observed that scaling the data quantity under a fixed data ratio results in performance conflicts or gains in the SFT (Source Fine-tuning) phase. To further investigate the role of data ratio in the data composition issue, we proposed two sub-research questions and conducted comparative experiments.

\textbf{Experimental Design:} 
% In real-world scenarios, we are particularly interested in providing a human-aligned model with excellent general instructional following ability and as-good-as-possible specialized ability. 
% We would not sacrifice the general ability for better specialized ability.
We consider coding and mathematics as a combined specialized data source, and the ShareGPT as the general data source. We designed three setups as follows which control the amount of one source of data and vary the ratio between general and specialized data.

% \textbf{1. Scaling general and specialized data:} 
% This is same as the experiment in RQ2, we sample \{1, 1/4, 1/16, 1/64, 1/256\} of training data together from GSM8k RFT, Code Aplaca, and ShareGPT as a mixture that controls the same data ratio and vary the data amount.

\textbf{1. Fixed general data, scaling specialized data:} We use a full training set of ShareGPT and sampled different proportions \{1, 1/4, 1/16, 1/64, 1/256\} of GSM8K RFT and Code Alpaca as a mixture.
% This setting controls the same data amount of general data and varies the ratio between general and specialized data.

\textbf{2. Fixed specialized data, scaling general data:} We use a full training set of GSM8K RFT and Code Alpaca and sample different proportions of ShareGPT as a mixture. 
% This setting controls the same data amount of specialized data and varies the ratio between general and specialized data.

\textbf{3. Fixed 1/64 general data, scaling specialized data}: Motivated by LIMA's setup \citep{lima}, we used a 1/64 ShareGPT set (about 1500 examples) and sampled different proportions of GSM8K RFT and Code Alpaca as a mixture. 
% This setting controls the same data amount of general data and 

\paragraph{Results and Analysis.} 
\textbf{Q1: Does the performance of the model vary with different ratios of general and specialized data?}
% 不同比例的通用数据与非通用数据会影响模型性能？
% 如figure3 中的(a)-(c) 所示,当我们采用固定general scaling code math (setup1)的设置时,随着数据量从1/256 scaling 至 1/1的情况下,MT-Bench socres并没有太大的性能波动(小于0.2),这说明非通用能力的data ratio变化并不会显著影响通用能力.除此以外,代码能力与数学能力的性能虽然有明显的波动,但其性能波动趋势与mix (code,math,general)设置基本一致,这说明general的数据比例并没有明显影响code math能力的正常scaling趋势.
% 同理在setup2,随着数据量从1/256 scaling 至 1/1的情况下,HumanEval与GSM8k均没有太大的性能波动,而MT-Bench也与Mix(code,math,general)的scaling curve相近,这也进一步印证了data ratio并不是数据组成问题的key factor,指导我们不用担心数据的长尾效应,而是更多关注于数据量的影响.
As illustrated in the top three graphs of Figure \ref{fig:data ratio}, we conduct ablation studies of the data ratio ($k$) between specialized and general abilities. 
To be noticed ratio is normalized by data amount, for example, $k=1$ means $\frac{\rm\ specialized\ use\ data\ amount}{\rm\ general\ use\ data\ amount} = \frac{\rm\ specialized\ all\ data\ amount}{\rm\ general\ all\ data\ amount}$.
We utilize a fixed specialized data setting (directly mixing 100\% code \& math data for training) and a fixed general data setting (100\% general data for training) as the baseline and observe:

\textbf{(1)} With the increase in the ratio of general data from 1/256 to 1/1, \textit{Fixed specialized data, scaling general data} setup exhibits similar performance to the setup that \textit{Fixed specialized abilities} in terms of math reasoning. This suggests that variations in the data ratio $k$ have minimal impact on math ability. 
We consider the reason that math and general abilities are non-conflict since they are too different in the semantic space.
However, when considering HumanEval, the \textit{Fixed specialized data, scaling general data} setup displays noticeable fluctuations compared to the baseline. 
% We attribute this to the significant differences in task formats and data distributions between math and general abilities, leading to a performance conflict caused by the data ratio. 
% Conversely, for code abilities, 
We attribute this to the inclusion of a certain proportion of code data in ShareGPT. Due to the differences in data format and distribution, the presence of similar data features exacerbates the performance conflicts between abilities when the data ratio $k$ increases. Further analysis of the distribution of different abilities is discussed in Section \ref{visual_41}. 

\textbf{(2)} With the increase in the ratio of specialized data from 1/256 to 1/1, the setup that \textit{Fixed general data, scaling specialized data} displayed no significant performance changes compared to the baseline. This echoes our hypothesis that when \textbf{there are significant differences in task formats and data distributions between different SFT abilities, the impact of data ratio is minimal}. 
However, when \textbf{there is some degree of similarities, the data ratio can lead to noticeable performance fluctuations.}

\textbf{Q2: Under extremely limited general data resources, does the ratio of specialized data have an impact on the model's performance?}
% 为了验证我们RQ3结论的泛化性,我们继续探寻在模型刚刚具备通用能力时(setup3),会不会受不同比例的非通用数据的影响.figure3(d)-(f)展示了setup1与setupt3的对比实验,我们发现无论通用能力的数据量是100%还是1/64,其MT-bench的性能并均没有随着非通用能力的数据比例变化而产生明显的波动.除此以外,在代码能力与数学能力上,setupt3呈现与setup1几乎相近的scaling趋势.这也进一步说明即使在极低通用数据量的情况下,模型也依旧对不同比例的非通用数据不敏感,这也进一步highlight数据比例并不是数据组成问题的key factor.
% To validate the generalization of our RQ3 findings, 
We further explore the impact of different ratios of specialized data when the model has just acquired a certain level of general human-aligning ability ($k=1/64$). The bottom 3 graphs of Figure \ref{fig:data ratio} present comparative experiments between two settings. We observe that regardless of whether the data amount for general capabilities is abundant ($k=1$) or scarce ($k=1/64$), the performance on MT-Bench shows no significant fluctuations with varying proportions of specialized data.  Furthermore, in mathematical reasoning, 1/64 general data setup exhibited a scaling trend that is almost identical to the full general data setup. 
However, for coding ability, with the same amount of code data and different ratios, code abilities are different in the two settings.
We still consider the reason is code data are partly related to ShareGPT data and cause the performance difference and provide an analysis in Discussion \ref{discusstion42}.

\subsection{RQ4. Performance Difference vs. Training Strategies}

% 1. 先介绍顺序,先混后顺序,以及直接混合的性能差异
% 2. 介绍data composition策略的提升---》达到一个平衡
%3. 介绍随着模型参数量增大的结论.

% In the context of real-world LLMs fine-tuning scenarios, carefully designed structure optimization methods often come with complex engineering issues. 
% Therefore, in this section, we aim to provide empirical guidance for mitigating performance conflicts among different data sources during the SFT phase through various simple and feasible training strategies, as outlined below:
We could feed these SFT datasets into models with different training strategies. In this section, We experiment with these settings and investigate how they influence each ability's performance.

\paragraph{Experimental Design:}Firstly, we introduce three kinds of naive training strategies as follows:

\textbf{1. Multi-task learning:} We directly mix different SFT data sources $D=\cup_{1\leq i \leq k}D_i$ and applying SFT. If we view each data source as a different task, this can be viewed as multi-task learning.

\textbf{2. Sequential Training:} We sequentially apply SFT on each dataset. Specifically, we sequentially trained on coding, math reasoning, and the general ability dataset.
Since the general ability is the most important one for human alignment, we put ShareGPT as our last dataset.

% This strategy begins by conducting fine-tuning on the LLaMA  model using the complete Code Alpaca dataset. Subsequently, the last checkpoint is loaded, and further training is conducted on the complete mathematical dataset. Finally, training is performed on the general human-aligning dataset. For a comprehensive and detailed discussion on the sequential strategy, please refer to the appendix.

\textbf{3. Mixed Sequential Training:} We apply multi-task learning on specialized datasets(code, math) first and apply SFT on the general ability dataset. These three approaches are presented in Figure \ref{fig:4_train_strategies}.
% This strategy entails the initial combination of the complete coding and mathematical datasets to form a mixed data source, which is subsequently used for fine-tuning the LLaMA  model. Then, we load the checkpoint and continue training it on the complete general human-aligning dataset. 

\paragraph{Results and Analysis:}
% Table1展现了LLaMA -7b在不同训练策略下的性能,我们相比于single source,直接混合策略,顺序训练在数学能力,代码生成上依旧维持了不错的性能,但是在通用能力表现并不佳.相反地,混合顺序训练虽然在非通用领域并没有展现出显著地能力,但是在通用能力上依旧维持了较为不错的性能.这也符合我们的预期,混合顺序训练在最后的finetuning阶段并没有受到非通用数据源的干扰,从而极大保留了通用能力,但是由于多阶段的训练导致它对之前的知识产生了灾难性遗忘,这也激励我们去探究如何在保证通用能力的情况下尽可能进一步缓解灾难性遗忘问题.

% In the case of 13B and 33B models, xxxxx

Table \ref{tab:main_result} presents performances under different training strategies in terms of mathematical reasoning, code generation, and general human-aligning ability.
Multi-task learning preserves specialized abilities among these strategies while hurting the general ability most among them.
Sequential training and mixed sequential training preserve general ability while losing too many specialized abilities.
% Multi-task learning and sequential training strategies maintain satisfactory performance, as opposed to the single source strategy. 
% Unfortunately, these strategies do not perform well on MT-Bench. Conversely, the mixed sequential training strategy, although it does not demonstrate significant enhancements in specialized domains, still maintains a relatively commendable level of performance in general capability. 
The observed outcome is in accordance with expectations, as during the final fine-tuning phase, the mixed sequential training strategy remains unaffected by specialized data, thereby effectively preserving its generalization capability. However, an inherent drawback of multi-stage training is the occurrence of catastrophic forgetting of prior knowledge, which motivates us to further explore methods that can alleviate catastrophic forgetting of specialized abilities while maximizing the preservation of general capability.
% Regarding larger models, it has been noted that multi-task training continues to pose the same issue, wherein the general capabilities of the model are significantly limited. However, sequential training and mixed sequential training have displayed impressive performance in terms of general capabilities; nevertheless, they have exhibited notable performance deficiencies and shortcomings in mathematical and coding abilities.
% \textbf{[need conclusion and analysis]}

\textbf{4. Dual-stage Mixed Fine-tuning (DMT):} 
Based on our observation from RQ1 to RQ4, we propose a new training strategy that can reduce the ability conflict during multi-task learning and relieve the issue of catastrophic forgetting during sequential training.
% We conducted refinements on the Mixed Sequential Training Strategy. 
From RQ1, the model needs large data amounts to activate specialized abilities.
From RQ2, multi-task learning with all amounts of specialized data and general data will hurt each ability.
From RQ3, a small amount of specialized data will not affect the general ability performance.
From RQ4, (mixed) sequential training forgets specialized abilities.
So the model needs to learn large amounts of specialized data and should not forget them during learning general ability.
A natural choice is to learn full amounts of specialized data first and add a small amount of specialized data to general data during the last stage of sequential training to prevent forgetting.
As shown in Figure \ref{fig:4_train_strategies}, we first apply SFT on the specialized dataset which is same as the first stage of the mixed sequential training strategy.
For the second stage, we perform SFT with a mixed data source comprising a combination of the general data and varying proportions $k$ ({1, 1/2, 1/4, 1/8, 1/16, 1/32}) of code and math data.
Adding code and math data in the second stage helps models to recall the specialized ability.
The results of DMT ($k=1/256$) are presented in Table \ref{tab:main_result} and the detailed scaling analysis of proportion $k$ can be found in the discussion.

% in the initial stage of the Sequential Fine-tuning, we incorporate a full collection of coding and mathematical data as mixed data source 1 in order to fine-tune the LLaMA  model.
% To mitigate the detrimental effects of catastrophic forgetting from the first stage, we introduce mixed data source 2 in the second stage, comprising a combination of general datasets and varying proportions k ({1, 1/2, 1/4, 1/8, 1/16, 1/32}) of code and math data. 
% The checkpoint from the first stage is then loaded, and the training procedure is continued using mixed data source 2.

% Model Accuracy vs. DMT Strategies.  在Table 1中可以看出,在大部分数据点上,随着融合非通用能力的比例减小,TMT策略下模型在GSM8k上的性能明显下降,而在MT-Bench性能逐渐提升.而当融合非通用能力的比例k=1/256时,对比naive training strategy中综合性能最好的Mixed Sequential Training Strategy,TMT在数学(40.48->46.47),code下都有一定的提升(15.24->15.85),这也意味着二阶段混合特定能力的数据对灾难性遗忘有显著的缓解作用,另人意外的是,在MT-Bench上,TMT(k=1/256)甚至有一定的提升,这证明TMT可以在保证通用能力的情况下尽可能进一步缓解灾难性遗忘问题.
% In the case of 13B and 33B models, 对比Mixed Sequential Training Strategy,TMT(k=1/256)对于数学推理(13b: 40.48->46.47),代码生成(13b: 18.3->19.5)的灾难性遗忘问题均有明显的缓解, 并且极大程度的保留了其通用能力(13b: 5.93->6.03),这也进一步验证了TMT的有效性

%---------------------------data compostion comparison 7B 13b 33b----------------------------------
\begin{table*}[t]
  \centering
  % \tiny
  \renewcommand{\arraystretch}{1.1}
  \resizebox{\textwidth}{!}{
  \begin{tabular}{lcccccccccc}
    \toprule
    \multirow{2}{*}{Methods}&\multicolumn{3}{c}{LLaMA -7B}&\multicolumn{3}{c}{LLaMA -13B}&\multicolumn{3}{c}{LLaMA -33B} \\
    \cmidrule(lr){2-4}
    \cmidrule(lr){5-7}
    \cmidrule(lr){8-10}
    & GSM8K & HumanEval & MT-Bench& GSM8K & HumanEval & MT-Bench& GSM8K & HumanEval & MT-Bench \\
    \midrule
    \multicolumn{1}{l}{\textit{Individual domain}} \\
    % \midrule
    General only& 11.10&	10.42& 5.88 &14.02 &16.40 &	 6.13 &26.06 &24.30 &6.63   \\
    Math only& 49.10 & 6.71 & 2.53& 51.40 & 12.8 & 2.54&57.91&15.5&3.18 \\
    Code only& 4.51& 18.40 & 4.30& 5.15& 17.1 & 3.53&6.06&26.82&4.18   \\

    \midrule
  
    \multicolumn{1}{l}{\textit{Different Training Strategies}} \\
     Multi-task learning &\textbf{47.53} & 14.63 &5.76 & \textbf{50.94 }  & \underline{19.50}	 & 5.73& \textbf{56.69} &18.9 &6.07   \\
    Sequential Training &31.39 &\underline{15.85}&  5.72 & 39.12  &\textbf{20.12}	  & \underline{5.93} &   47.27   &\underline{24.80} & \textbf{6.73}   \\
    Mixed Sequential Training & 32.60   &15.24	  & \underline{6.02}& 40.48 & 18.30 & \underline{5.93}& 44.24  &24.4   & 6.43 \\
    
    DMT(k=1/256) & \underline{41.92}& \textbf{17.68} \ & \textbf{6.08} & \underline{46.47}  &\underline{19.50} &  \textbf{6.03} & \underline{56.36}  &\textbf{25.00}   & \textbf{6.73}    \\
    \bottomrule
  
  \end{tabular}}
  \caption{The results of LLaMA-7B, 13B, 33B under different training strategies on three benchmarks. The top two results across different strategies are marked with \textbf{bold} and \underline{underlined}.}
  \label{tab:main_result}
  \vspace{-0.2cm}
  
\end{table*}
%---------------------------data compostion comparison 7B 13b 33b----------------------------------

\textbf{Model Accuracy vs. DMT Strategies.} In Table \ref{tab:main_result}, LLaMA-7B with DMT ($k=1/256$) strategy perform significant improvement in mathematical reasoning (32.6 to 41.92) and code generation (15.24 to 17.68) compared to the mixed sequential training strategy, which indicates a significant alleviating effect of mixing specialized capability data in the last fine-tuning stage on catastrophic forgetting. Surprisingly, DMT ($k=1/256$) even exhibits a slight improvement on MT-Bench, further highlighting its ability to alleviate catastrophic forgetting while effectively preserving general capability.

Regarding the 13B and 33B models, DMT ($k=1/256$) demonstrates noticeable alleviation of catastrophic forgetting in mathematical reasoning (13B: 40.48 to 46.47 / 33B: 44.24 to 56.36) and code generation (13B: 18.3 to 19.5 / 33B: 24.4 to 25.5) compared to the mixed sequential training strategy. Additionally, it significantly retains its general capability (13B: 5.93 to 6.03 / 33B 6.43 to 6.69). 
% When the model is trained with single domain, 
% \textbf{[need some insight analysis]}. 
Therefore, these results serve as additional validation of the efficacy of DMT in mitigating catastrophic forgetting while maintaining general capability \footnote{To verify the effectiveness of DMT strategy on relatively OOD benchmarks, we further evaluate it on MBPP and MATH in Appendix \ref{sec:app_f}.}.

%----------------tsne---------------------
% As the proportion of specialized capabilities decreases, the model's performance on GSM8k exhibits a significant decline, while gradually improving on MT-Bench. Notably, when the proportion of non-general capabilities is k=1/256,
\begin{figure*}[h]
\centering
\begin{subfigure}[t]{0.23\linewidth}
\centering
\includegraphics[width=1.1\linewidth]{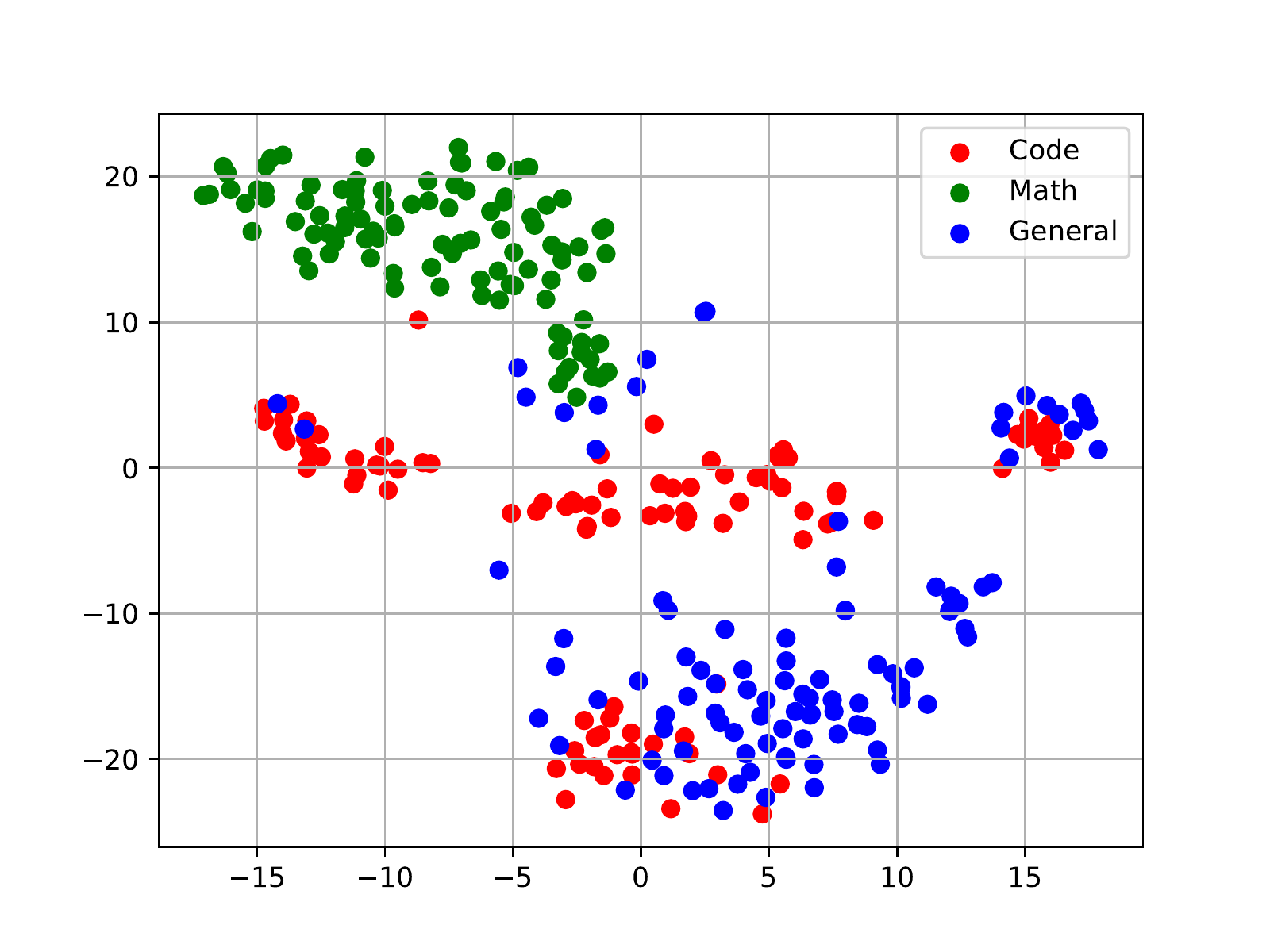}
\caption{LLaMA-13B}
\label{fig:tsne_base13b}
\end{subfigure}%
\begin{subfigure}[t]{0.23\linewidth}
\centering
\includegraphics[width=1.1\linewidth]{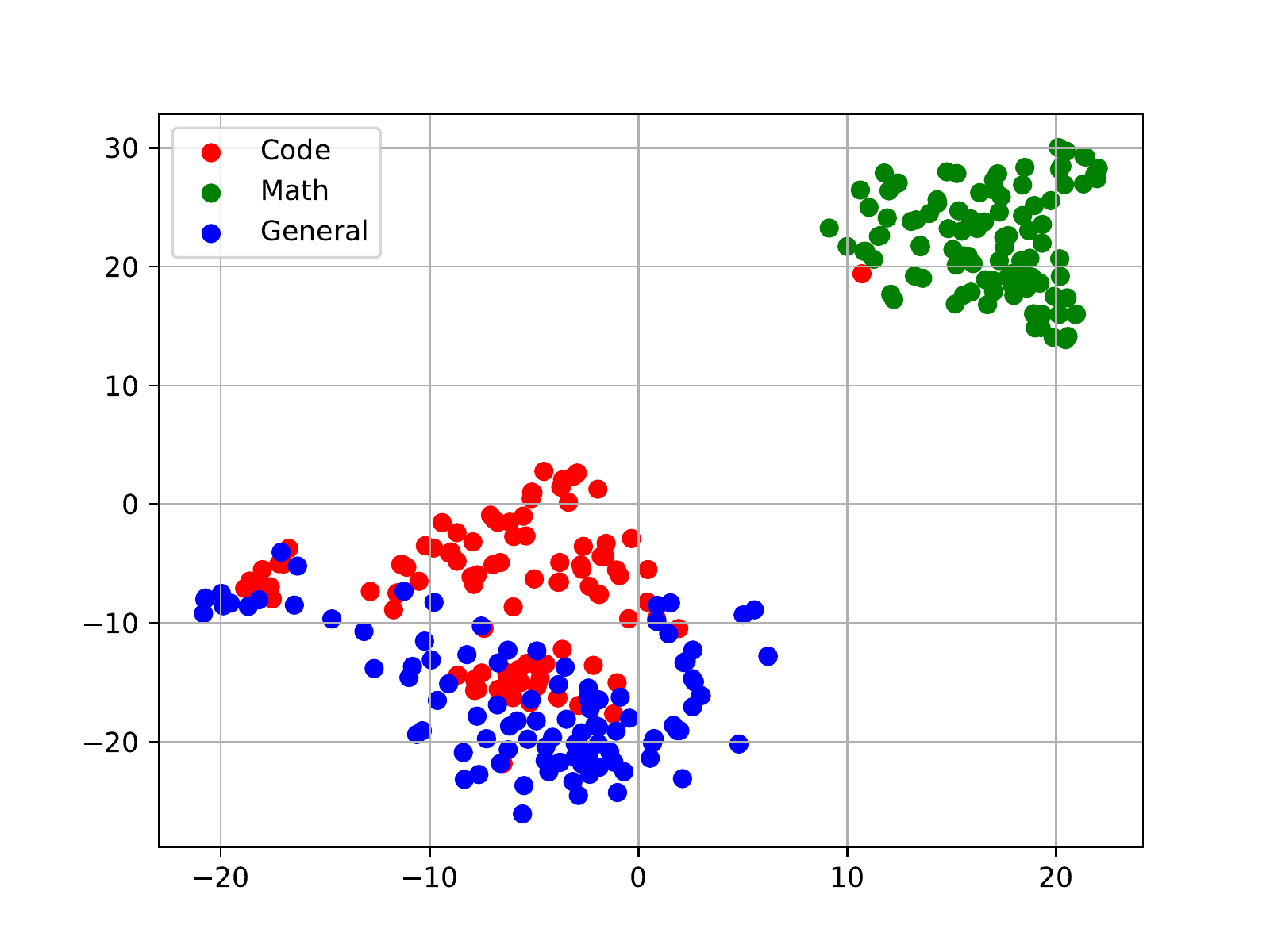}
\caption{LLaMA-13B with DMT}
\label{fig:tsne_dmt13b}
\end{subfigure}%
\begin{subfigure}[t]{0.23\linewidth}
\centering
\includegraphics[width=\linewidth]{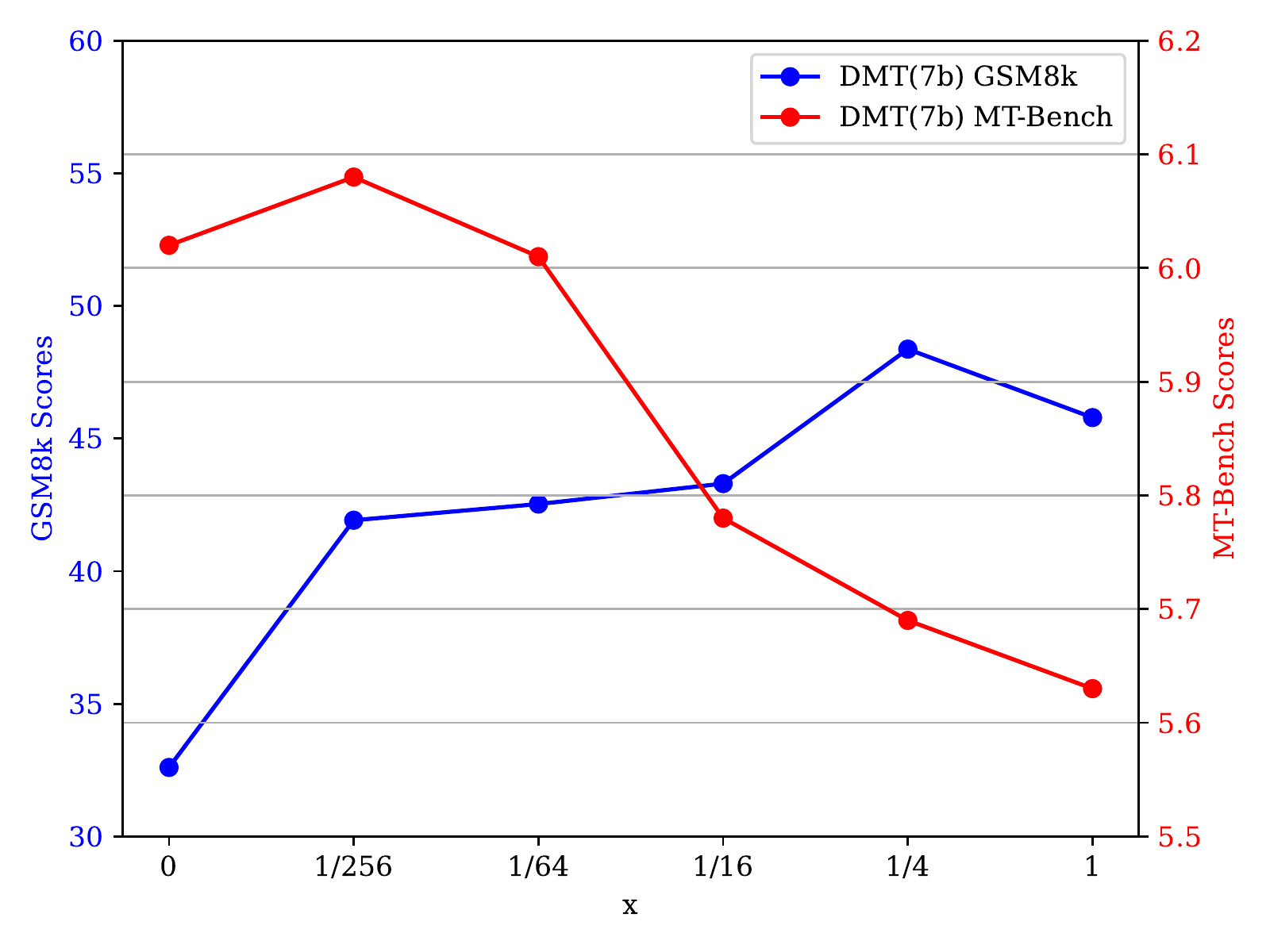}
\caption{DMT scaling (7B)}
\label{fig:dmt_scaling_7b}
\end{subfigure}%
\begin{subfigure}[t]{0.23\linewidth}
\centering
\includegraphics[width=\linewidth]{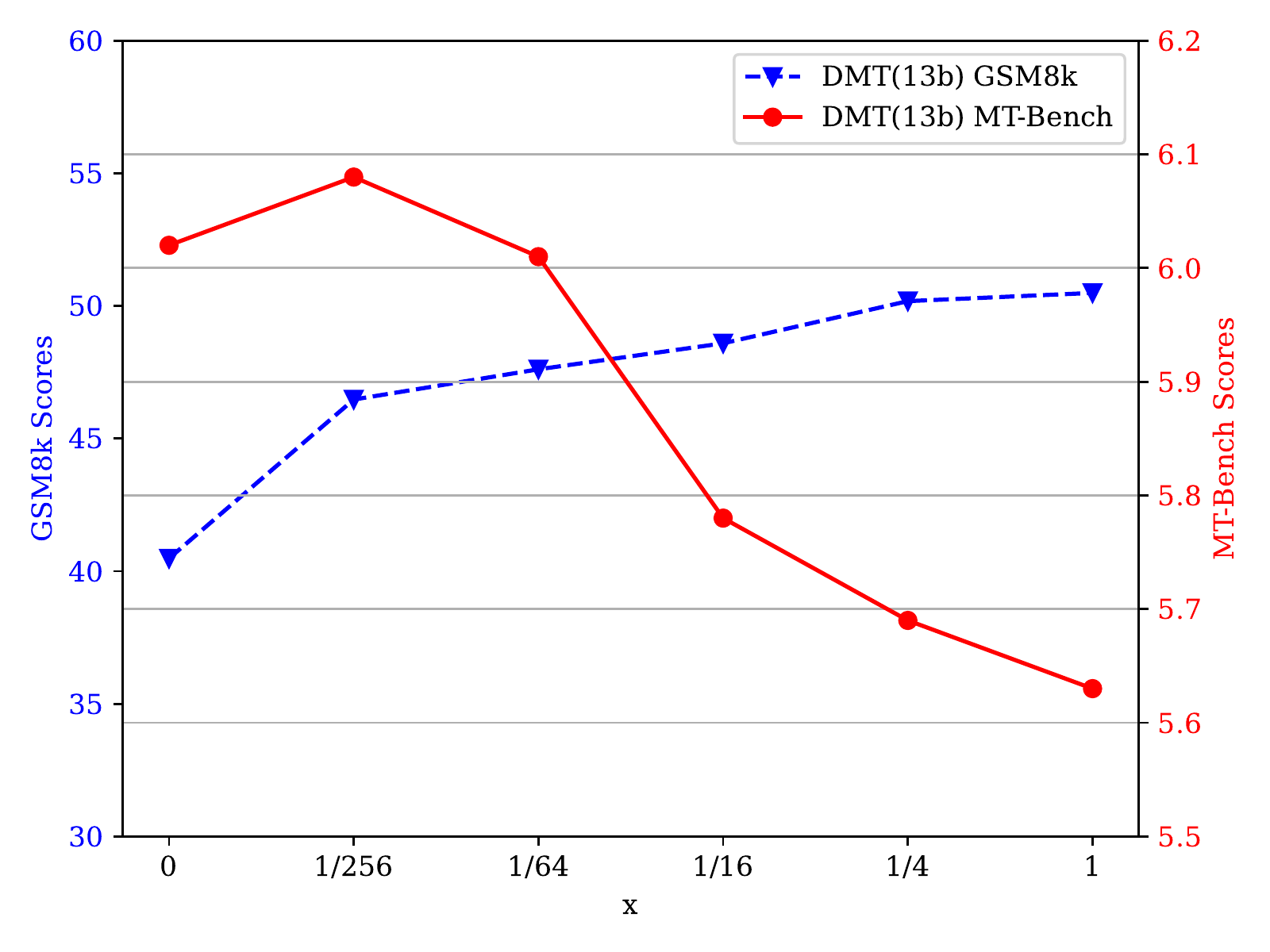}
\caption{DMT scaling (13B)}
\label{fig:dmt_scaling_13b}
\end{subfigure}
\caption{The left two figures show the t-SNE plots of LLaMA-13B and LLaMA-13B with the DMT strategy. The two right figures show the performance scaling of LLaMA-7B \& 13B with DMT under different $k$ values.}
\label{fig:tsne_dmt_scaling}
\end{figure*}

% %----------------tsne---dmt scaling---------------------
% \begin{figure*}[h]
% \centering
% \begin{subfigure}[t]{0.23\linewidth}
% \centering
% \includegraphics[width=1.2in]{figures/tsne_plot_base13b.pdf}
% % \caption{1}
% \end{subfigure}%
% \begin{subfigure}[t]{0.23\linewidth}
% \centering
% \includegraphics[width=1.2in]{figures/tsne_plot_dmt13b.pdf}
% % \caption{2}
% \end{subfigure}%

% \begin{subfigure}[t]{0.23\linewidth}
% \centering
% \includegraphics[width=1.2in]{figures/dmt_scaling_7b.pdf}
% \end{subfigure}%
% \begin{subfigure}[t]{0.23\linewidth}
% \centering
% \includegraphics[width=1.2in]{figures/dmt_scaling_13b.pdf}
% % \caption{3}
% \end{subfigure}
% \caption{The left two figures show the t-SNE of LLaMA-13B and LLaMA-13B with DMT(k=1/256) stategy. The right figure shows performances of LLaMA-13B with DMT under different $k$.}
% \label{fig:dicuss1_2}
% \end{figure*}

\section{Discussion}
\subsection{Visualization of Different SFT Abilities}
\label{visual_41}
In the aforementioned analysis of data composition, we observed a significant performance degradation when different data sources are directly mixed. In this section, our aim is to explore the potential mutual influence of semantic representation distributions among different data sources.
Specifically, we randomly sampled 100 queries from CodeAlpaca, GSM8k RFT, and ShareGPT datasets and extracted the hidden layer representations located in the Middle layer (15th) of the model. Subsequently, we employed the t-SNE toolkit \cite{van2008visualizing} to visualize the representations of the three types of capabilities. The results in Figure \ref{fig:tsne_dmt_scaling} illustrate a notable collapse phenomenon in the semantic representations of both the original LLaMA-13b and LLaMA-13b with DMT (k=1/256). While both models exhibit a certain level of separation in the mathematical data representations, there remains a certain degree of overlap between the representations of code and general samples. In Appendix \ref{sec:app_g}, we further discuss the visualization of semantic spaces at different layers of LLaMA 7B \& 13B.
% This observation suggests a substantial similarity in the distribution patterns of these two types of data, which further poses a challenge for the future optimization of these two domains in SFT phrase.

\subsection{Ablation of the Specialized Domains in ShareGPT}
\label{discusstion42}
In RQ2, we observe using mixed data sources resulted in improved abilities under low-resource conditions but diminished abilities under high-resource conditions when compared to single data sources. However, the presence of coding and mathematical samples within the ShareGPT introduces uncertainty regarding whether the performance gain under low resources is solely attributed to these specific coding \& mathematical data or other orthogonal samples in the general dataset (e.g., translation or extraction). Hence, the objective of this section is to investigate whether the conclusions drawn in Section 3.3 remain valid after removing the code and math samples within ShareGPT.

\textbf{Experimental Design:} We employed an open-set tagger InsTag \citep{instag} to annotate samples in ShareGPT. To filter out data related to coding and mathematical abilities, we conduct regular expression matching to eliminate instances where the tags contain keywords ``code'' or ``math''. Finally, we obtain a ShareGPT dataset devoid of any code or math-related information (reducing from 86K to 63K). In alignment with the settings in Section 3.3, we sampled different proportions of training data ({1, 1/4, 1/16, 1/64, 1/256}) from GSM8K, Code Alpaca, and the modified ShareGPT dataset (without code math). These samples were directly mixed according to the corresponding proportions. Subsequently, the LLaMA models were fine-tuned by using this mixed dataset.

\begin{figure}[t]
    \centering
    \small    \includegraphics[width=\linewidth]{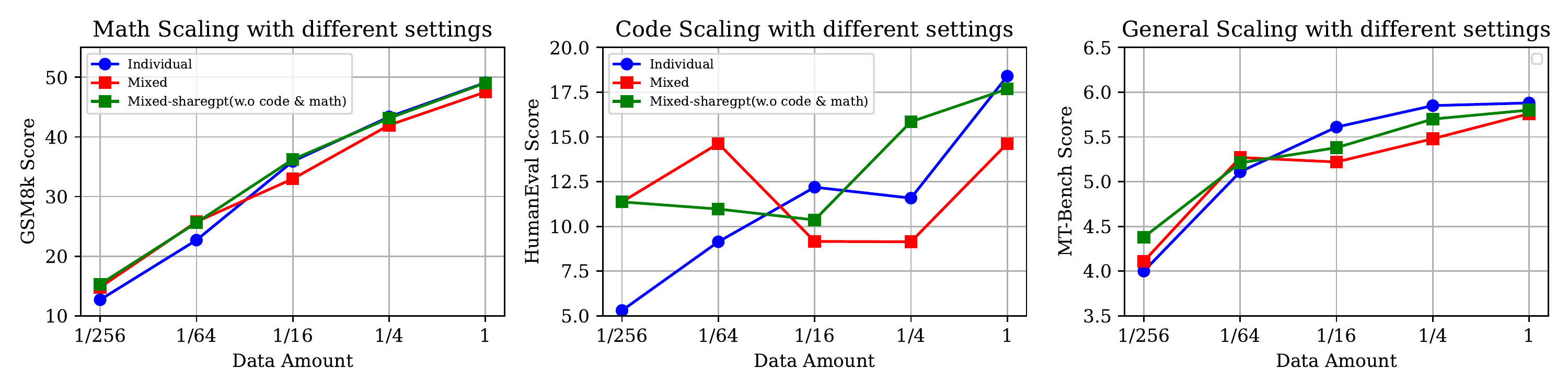}
    \caption{The scaling curve after ablating code and math-related samples from ShareGPT.}
    \label{fig:instag}
    \vspace{-0.2cm}
\end{figure}

\textbf{Results and Analysis.} Figure \ref{fig:instag} shows the results of our experiment. Removing the code and math from ShareGPT not only mitigates the performance conflicts among different abilities to some extent under high-resource conditions but also maintains stable gains in low-resource settings. We propose that the potential reason behind these findings lies in the differences in the distribution of code and math data between ShareGPT, CodeAlpaca, and GSM8K RFT datasets. This distribution gap introduces an extra noise during the SFT phrase, while its removal enables the model to better generalize coding and mathematical abilities. Furthermore, in low-resource scenarios, this phenomenon indicates that the code and math samples in ShareGPT are not the key factor contributing to performance improvements, but rather the diversity and variability of the data \citep{longpre2023flan}. 
In summary, the presence of code math data within ShareGPT does not emerge as a key factor impacting the performance gains identified in Section 3.3, highlighting the generalization of our conclusions.

%-----------instag code math------------
%-----------instag code math------------

\subsection{Specialized Data Amount in DMT}
% 为了gain an insight into RQ3.5中DMT的比例系数k取值的影响,我们以scaling的视角进一步探究了DMT不同k的取值,在特定能力与通用instruction following能力上的关系【脚注, DMT(13b)关于k的scaling讨论在附录中,由于code的scaling曲线相对irralvent,我们没有放上】.可以明显看出当我们的DMT(k=0)【DMT(k=0)意味着他等价于Mix sequential training设置】调整到k=1/256时,模型在特定能力与通用能力上均有明显的提升,而当k的取值从1/4增加到1时,模型的特定能力与通用能力上均有一致性的下降,我们认为这也与RQ1中高资源互相冲突,低资源互相增益的结论相得益彰,印证了DMT策略的优势. 

% 除此以外,随着k从1/256增长到1/4时,我们观察到通用能力与GSM8k能力基本呈现出一个线性的此消彼长趋势,这也提示了我们需要根据特定的训练需求来决定k的取值,以达到通用能力与特定能力的平衡性.如果需要训练通用大模型,DMT的k取值可以在(0,1/64)之间,而如果我们要训练垂直领域大模型,k的取值区间推荐在(1/16,1/4),这为SFT模型训练提供了一定经验性的方案.
We investigate how different values of $k$ influence model performance and results shown in Figure~\ref{fig:tsne_dmt_scaling}.
When we adjust $k$ from 0 to 1/256 ($k=0$ is equal to mixed sequential training), the SFT models show significant improvements in both specialized ability and general human-aligning ability. On the contrary, as $k$ increased from 1/4 to 1, the model exhibited a decline in general ability. We believe this is in line with the findings in RQ2, which concluded that high-resource settings lead to conflicts while low-resource settings lead to gains in mixed sources.
% , thus confirming the superiority of the DMT strategy compared to other naive training strategies.
Furthermore, as $k$ increased from 1/256 to 1/4, we observe a linear inverse trend between general ability and specialized ability, especially an increase in general ability coincided with a decrease in specialized ability. This suggests $k$ needs to be tuned based on specific requirements in order to achieve a balance between multiple abilities.

% \begin{figure}[t]
%     \vspace{-0.3cm}
%     \centering
%     \subfigure[PSSAT]{
%         \includegraphics[width=.5\textwidth]{figures/tsne_plot_base7B.pdf}
%         \label{noise}
%     }
%     \subfigure[Noise-BERT]{
% 	\includegraphics[width=.5\textwidth]{figures/tsne_plot_best.pdf}
%         \label{noise2}
%     }
%     \vspace{-0.3cm}
%     \caption{The t-SNE visualization of the LLaMA -7B baseline and the best data composition strategy. } 
%     \label{fig:visual}
%     \vspace{-0.4cm}
% \end{figure}

\section{Conclusion}
We explore the data composition in the SFT phase, focusing on mathematical reasoning, code generation, and general human-aligning abilities. We formulate four research questions to guide our investigation and analyze the scaling trends between different abilities and factors (e.g. data amount, data ratio, model parameters, and training strategies). Our findings reveal distinct scaling patterns among different abilities, with larger models demonstrating superior performance when trained with the same amount of data. Moreover, mixing data sources in the SFT phase improves performance in low-resource scenarios but diminishes in high-resource scenarios. Interestingly, the phenomenon of low-resource gain becomes more prominent as the model parameter size increases. Furthermore, our observations indicate that data amount directly influences performance conflicts, whereas the impact of data ratio is insignificant within our experimental setup. Finally, regarding the SFT strategies, we demonstrate our proposed DMT strategy effectively alleviates performance conflicts, offering a promising solution to activate multiple abilities.

\section*{Limitations}
Due to our use of the large language model LLaMA-33B, the extensive computational resources and time required for both training and inference may limit its applicability. The datasets used in this article are all open source, so there are no ethical or moral issues; However, inappropriate prompts and noisy training corpora can potentially lead to privacy and bias issues with LLMs. Furthermore, the evaluation benchmark MT-Bench relies on GPT-4 for scoring, which may result in some variability in the results, and these may not always align perfectly with human judgment standards. In this paper, we primarily focus on three SFT capabilities that are of great interest in the LLMs community, including mathematical reasoning, code generation, and general human-aligned ability. To verify the generality of our conclusions, we further explore three additional SFT capabilities in the appendix. Nevertheless, there are still many other SFT capabilities (such as creative generation) within the LLMs community that have data composition issues waiting to be explored by researchers, which will also be the focus of our future research efforts.

\bibliography{anthology,custom}

\appendix

\newpage
\section{SFT Datasets}
\label{sec:app_a}
We investigate the data composition issues of mathematical reasoning, coding, and general capabilities in the SFT stage from the following SFT datasets.

\begin{itemize}
\item \textbf{Code Alpaca} \citep{codealpaca} aims to build and share an instruction-following LLaMA  model for code generation. which is fully based on Stanford Alpaca and contains 20K data used for fine-tuning the model. The Code Alpaca dataset has been open-sourced\footnote{\url{https://github.com/sahil280114/codealpaca}}.

\item \textbf{GSM8K RFT} \citep{rft} is a mathematical dataset enhanced by integrating multiple reasoning paths based on the original GSM8K dataset \citep{gsm8k} through the rejection sampling. It contains 7.5K questions and 110K responses in the training set. The GSM8k RFT dataset has been open-sourced \footnote{\url{https://github.com/OFA-Sys/gsm8k-ScRel}}.

\item \textbf{ShareGPT} refers to the multi-turn chatting histories used by Vicuna \cite{vicuna2023}. ShareGPT includes 86K human queries and responses from ChatGPT and other chatbots. The GSM8k RFT dataset has been open-sourced \footnote{Exact dataset of ShareGPT (\url{https://sharegpt.com/}) has not been released. We instead
use a reproduced version from \url{https://huggingface.co/datasets/anon8231489123/ShareGPT_Vicuna_unfiltered} cleaned raw dataset, and follow Vicuna preprocess.}.
\end{itemize}

The following table \ref{tab:app_stastic} presents the statistics of three datasets at different subset proportion (k).

%------------------------------7B----------------------------------

\begin{table}[h]
    \centering
    \small
    \renewcommand{\arraystretch}{1.2}
    \begin{tabular}{l|ccc}
  \hline
   Data Ratio & GSM8K RFT & CodeAlpaca & ShareGPT\\
  \hline

 % \hline
  K=1/1 & 110142 &20022 &86060 \\
  K=1/4 & 27535 & 5005 & 21515 \\
  K=1/16 & 6883 &1251 & 5378\\
  K=1/64 & 1720 & 312 & 1344\\
  K=1/256 & 430 & 78 & 336 \\

\hline
\end{tabular}
\caption{Data statistics of three datasets at different subset proportion (k).}
\label{tab:app_stastic}
\end{table}

%------------------7B--------------------------

%110142
%86060
%20022

\section{Evaluation metrics}
\label{sec:app_b}
We use the following metrics to measure the aligned large language models.

\begin{itemize}
\item \textbf{HumanEval} \citep{chen2021codex} consists of 164 original programming problems, with an average of 9.6 test cases allocated to each problem. To ensure a thorough assessment of the functional correctness of LLM-synthesized code, HumanEval+ extends the number of test cases significantly, averaging at 774.8 test cases per problem. We use the same method as \cite{chen2021codex}to obtain unbiased estimates of Pass@k under greedy decoding. To facilitate the reproducibility of our results, we use the open-source github repository BigCode \citep{bigcode-evaluation-harness} to evaluate all the HumanEval scores in this paper \footnote{\url{https://github.com/bigcode-project/bigcode-evaluation-harness}}.

\item \textbf{GSM8K} \citep{gsm8k} is a math word problem dataset used to measure large language model math reasoning ability. We use the default test set to measure the model. We calculate the score based on greedy decoding accuracy (maj@1). In this paper, we use the open-source github repository gsm8k-ScRel \footnote{\url{https://github.com/OFA-Sys/gsm8k-ScRel}} to evaluate all the GSM8k scores.

\item \textbf{MT-Bench} \citep{zheng2023judging}  is a significant benchmark that contribute to the evaluation and advancement of chatbot models and LLMs in different contexts. MT-Bench\footnote{\url{https://huggingface.co/spaces/lmsys/mt-bench}} evaluates LLMs on multi-turn dialogues using comprehensive questions tailored to handling conversations. It provides a comprehensive set of questions specifically designed for assessing the capabilities of models in handling multi-turn dialogues.

\end{itemize}

We also supplement more benchmark evaluation results in the appendix F to verify the generalization of our conclusions:
\begin{itemize}
\item \textbf{MATH} \citep{hendrycks2021measuring} is a dataset with challenging high-school math problems. Problems are classified into the following topics: Prealgebra, Algebra, Number Theory, Counting and Probability, Geometry, Intermediate Algebra, and Precalculus. Problems in MATH are harder and more diverse than in GSM8K. In this paper, we use the open-source github repository gsm8k-ScRel to evaluate all the MATH scores.
% We train on 7,500 training samples from the original MATH dataset and test on 500 testing samples from \citep{lightman2023lets} for quick evaluation.
We use 500 test problems from \cite{lightman2023lets} as out-of-domain math benchmark.

\item \textbf{MBPP} \citep{austin2021program} consists of around 1,000 crowd-sourced Python programming problems, designed to be solvable by entry-level programmers, covering programming fundamentals, standard library functionality, and so on. Each problem consists of a task description, code solution and 3 automated test cases. To facilitate the reproducibility of our results, we use the open-source github repository BigCode \citep{bigcode-evaluation-harness} to evaluate all the MBPP scores in this paper.

\end{itemize}

%-----------single scaling------------
\begin{figure*}[t]
    \centering
    \small    \includegraphics[width=\linewidth]{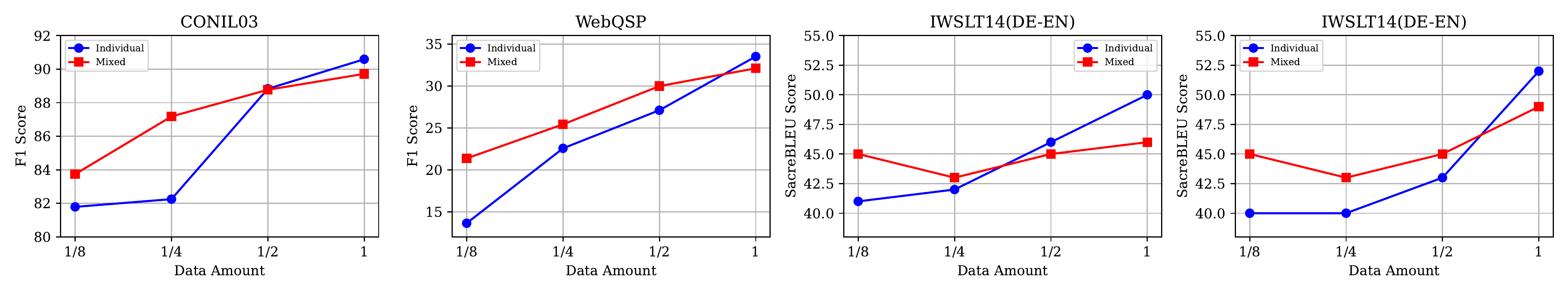}
    \caption{The scaling curve of LLaMA-7B in WebQSP, CoNLL 2003, IWSLT14(de-en), IWSLT14(de-en).}
    \label{fig:app_more_scaling}
\end{figure*}
%-----------single scaling------------

\section{Implementation Details}
\label{sec:app_c}
We fine-tune all the SFT datasets with 3 epochs and a batch size of 128 on NVIDIA A100 GPUs. We use 8 GPUs for 7B and 13B models, 16 GPUs for 33B models during fine-tuning. We use a peak learning rate of 2e-5 with a 3\% learning rate warmup. We evaluate the results on the final epoch. We use greedy decode to calculate Pass@1 and maj@1. Since the scores of MT-bench will fluctuate, we conducted three experiments and took the average. 

All experiments are conducted using the default template of the FastChat framework \citep{zheng2023judging}, as shown in the figure below:

\begin{tcolorbox}[
colback=white!10!white,
colframe=black!75!black,
title=Prompt Template,
breakable]
\label{response-aug prompt}
A chat between a curious user and an artificial intelligence assistant. The assistant gives helpful, detailed, and polite answers to the user's questions. \textbf{USER:} \{Query\} \textbf{ASSISTANT:}
\end{tcolorbox}

To facilitate the replication of our results, all datasets and evaluation benchmarks used in our experiments have been open-sourced and their detailed sources are indicated. We will also open-source our code after the blind review process.

\section{Estimating FLOPs of SFT}
\label{sec:app_d}

\paragraph{Training FLOPs.} We mainly follow the notations of \citep{kaplan2020scaling} here. 

For each input sample of length $n_{ctx}$ in SFT dataset (GSM8K, CodeAlpaca, ShareGPT), we can split it into two parts:
\begin{equation}
n_{ctx} = n_{Q}+n_{R}
\end{equation}

\begin{equation}
    C_\text{train} \approx 6N n_{ctx} N_{s}
\end{equation}

where $n_{Q}, n_{R} $ denotes the length of question and generated answers respectively. $N$,$N_{s}$ denotes the non-embedding parameters and the numbers of samples.

Therefore, We estimate the SFT FLOPs following \citep{kaplan2020scaling} and GPU times in Table~\ref{tab:flops}.

%------------------------------7B----------------------------------

\begin{table}[t]
       \centering
    \small
    \renewcommand{\arraystretch}{1.2}
    \begin{tabular}{l|cccc}
  \hline
  Model size & 7B & 13B & 33B \\

 %  \hline
 % 1 Mix(code,math,general)&47.53 & 14.63  &5.76       \\
 % 1/4 Mix(code,math,general)& 41.98 & 9.14  & 5.48    \\
 %  1/16 Mix(code,math,general)& 32.97 & 9.16  & 5.22     \\
 % 1/64 Mix(code,math,general)& 25.77 &14.63   & 5.27    \\
 % 1/256 Mix(code,math,general)& 14.78 & 11.37   & 4.11 \\

%[11.37, 14.63, 9.16, 9.14, 14.63]
\midrule
\multicolumn{1}{l}{\textit{GSM8k RFT}} \\
\midrule
  SFT FLOPs& $2.4\times10^{18}$ & $4.3\times10^{18}$	 & $1.1\times10^{19}$ \\
  SFT GPI hrs& 6.1   &12.1	 & 37.4  \\

\midrule
\multicolumn{1}{l}{\textit{Code Alpaca}} \\
\midrule

 SFT FLOPs& $4.7\times10^{17}$ & $7.8\times10^{17}$	 & $2.0\times10^{18}$  \\
  SFT GPI hrs&1.2   &2.5	 & 8.2 \\

\midrule
\multicolumn{1}{l}{\textit{ShareGPT}} \\
\midrule

 SFT FLOPs& $2.2\times10^{18}$ & $3.9\times10^{18}$	 & $9.7\times10^{19}$ \\
  SFT GPI hrs&5.4   &10.9	 & 34.0 \\

\hline
\end{tabular}
\caption{The statistics of FLOPs and GPU hours required for SFT. For 33B, we use DeepSpeed ZeRO3 \citep{deepspeed} for distributed training. All the GPU hours are based on NVIDIA A100 80GB GPU. Note we use non-embedding parameters to compute FLOPs in our experiments.}
\label{tab:flops}
\end{table}

%------------------7B--------------------------

% \paragraph{Inference FLOPs}

% We roughly computed the FLOPs of each token during the forward pass:
% \begin{equation}
%     C_\text{forward}(n_\text{ctx}) = 2N + 2n_\text{layer}n_\text{ctx}d_\text{model}
% \end{equation}

% To ensure the results were more accurate and reliable, we also took into account the Key-Value (KV) cache during the decoding procedure.

% \begin{equation}
% KV_\text{cache} \approx 4 n_\text{layer}d_\text{model}^2 
% \end{equation}

% Therefore,  we obtain the FLOPs per token during the forward pass considering the KV cache.
% \begin{align}
%     C_\text{forward}^{'}(n_{ctx}) & = 2N + 2n_\text{layer}n_{ctx}d_\text{model} - KV_\text{cache} \\
%     & =24n_\text{layer}d_\text{model}^2 + 2n_\text{layer}n_{ctx}d_\text{model} - 4 n_\text{layer}d_\text{model}^2 \\
%     & =20n_\text{layer}d_\text{model}^2+ 2n_\text{layer}n_{ctx}d_\text{model}\\
%     & \approx 1.66N + 2n_\text{layer}n_{ctx}d_\text{model} 
% \end{align}

% The total inference FLOPs are computed as follows: 
% \begin{equation}
%     C_\text{total} = N_{s} \cdot [ n_{q} C_\text{forward} (n_{q}) + \sum_{i=n_{q}}^{n_{q}+n_{r}} i \cdot  C_\text{forward}^{'}(i) ]
% \end{equation}
% where $N_{s}$ denotes the numbers of samples. $n_{q}, n_{r} $ denotes the average length (tokens) of the user query and generated response respectively. In GSM8K dataset,  
% $n_{q}\approx 66$ and $n_{r} \approx 130 $.

\section{Validation Experiments in More SFT Abilities}
\label{sec:app_e}
To validate the generalization of our conclusions, we selected representative datasets to evaluate the capabilities of large models across different dimensions. These dimensions include \textbf{World Knowledge} : WebQuestionsSP \citep{yih-etal-2016-value}, \textbf{Language Understanding}: CoNLL 2003 \citep{tjong-kim-sang-de-meulder-2003-introduction}, and \textbf{Translation}: IWSLT14 \citep{cettolo-etal-2014-report}

\paragraph{Experimental Design:} Align the settings of RQ1 and RQ2, we introduce two settings as follows:

\textbf{1. Individual Domain:} We conduct SFT on LLaMA of various sizes using \{1, 1/2, 1/4, 1/8\} proportions \footnote{Because these three datasets have relatively small amounts of data (a few thousand), the scaling range is from 1/1 of the data volume to 1/8 of the data volume.} of the training set obtained from WebQSP, CoNLL 2003, and IWSLT14 seperately. This allowed us to evaluate each ability with various data sizes and model sizes.

\textbf{2. Mixed Domain:} We sampled \{1, 1/2, 1/4, 1/8\} amounts of training data from WebQSP, CoNLL 2003, and IWSLT14, and directly mixed them according to the corresponding proportions. In this way, we constructed datasets with fixed proportions of different ability domains, while varying the total data amount. These datasets are then used for fine-tuning the LLaMA models.

\begin{table*}[h]
	\centering
	\footnotesize
	\begin{tabular}{c|c|c|c|c|c|c|c}	
	\toprule
	\multirow{2}{*}{Datasets} & \multicolumn{3}{c|}{CONIL03}  & \multicolumn{2}{c|}{WebQSP} & \multicolumn{2}{c}{IWSLT14} \\

    & P & R & F1 & F1 & Hits@1 &de-en &en-de  \\
    \midrule
    Single Domain(1/1)  &91.89 &89.33  &90.59  &33.51   &64.12   & 50   &52      \\ 	
    Single Domain(1/2) &90.59 &87.15 &88.83 & 27.10   &61.87    &46  &43     \\ 	
    Single Domain(1/4)  &85.24 &79.46 & 82.25   &22.56   & 61.38   &42  &40    \\ 	
    Single Domain(1/8)  &83.22 &80.42  &81.79  &13.63  &49.05    &41  &40   \\ 	

    \midrule
    Mixed Domains(1/1)  &91.74&87.79 &89.72   &32.10   &63.70    &46  &49      \\ 	
    Mixed Domains(1/2)  &90.69& 86.93 &88.77  &29.98   &62.29  &45  &45      \\ 	
    Mixed Domains(1/4)  &88.81 &85.62  &87.18    &25.42   & 58.02   &43  &43     \\ 	
    Mixed Domains(1/8)   &86.47 &81.18  &83.74   &21.36   &56.86    &45  &45   \\

		\bottomrule
	\end{tabular}
  \caption{Results in other domains for single and mixed source settings based on LLaMA-7B.} 
  \label{tab:other_datasets}	
\end{table*}

\textbf{Analysis.} As shown in Figure \ref{fig:app_more_scaling} and Table \ref{tab:other_datasets}, we have following observations.

For the \textbf{individual domain}, the performance (P, R, F1) of the model in the language understanding (NER) task shows a positive correlation with the scaling curve of data volume. These two abilities exhibit similar scaling curve trends as the mathematical ability performance in RQ1. In the case of world knowledge (WebQSP), a similar positive correlation trend is observed in terms of F1 and Hits@1. However, when the data ratio is reduced from 1/4 to 1/8, there is a significant performance fluctuation, particularly in the performance of translation ability, which shows a relatively irregular trend. These conclusions further support the core conclusion of RQ1 that different data exhibit different scaling curves.

For the \textbf{mixed domains}, the findings align with the conclusions in RQ2, where abilities are improved with low-resource and decreased with high-resource compared to individual source abilities. This consistent conclusion holds for world knowledge, language understanding, and translation abilities.

\section{Results on OOD Benchmarks in Math and Code}
\label{sec:app_f}
To validate the generalization of our findings on other benchmarks, we utilized GSM8K and Code Alpaca as the training sets. We further evaluated the results on the individual domain, mixed domain, and different training strategies on other specialized ability benchmark, including MATH and MBPP, which is illustrated in Table \ref{tab:other_benchmark} and Figure \ref{fig:app_ood_scaling}. We have the following findings:

%-----------single scaling------------
\begin{figure}[t]
    \centering
    \small    \includegraphics[width=\linewidth]{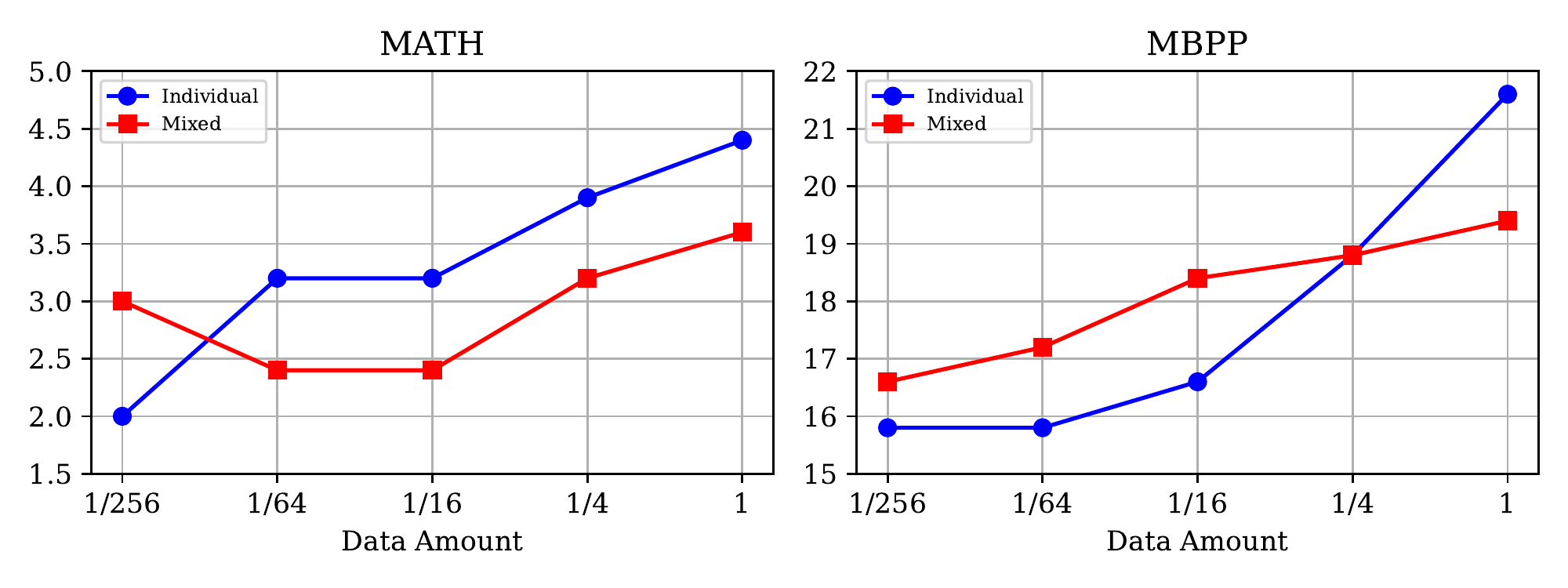}
    \caption{The scaling curve of LLaMA-7B on MATH and MBPP benchmarks.}
    \label{fig:app_ood_scaling}
\end{figure}
%-----------single scaling------------

\begin{table*}[h]
    \centering
  \small
  \renewcommand{\arraystretch}{1.1}
      {
  \begin{tabular}{l|cc|cc}
    \toprule
    \multirow{2}{*}{Methods}&\multicolumn{2}{c|}{Math Benchmarks}&\multicolumn{2}{c}{Code Benchmarks} \\
    \cmidrule(lr){2-3}
    \cmidrule(lr){4-5}
   
    & GSM8K & MATH & HumanEval& MBPP \\
    \midrule

    \multicolumn{1}{l}{\textit{Individual domain (Scaling)}} \\
    \midrule

   Single Domain(k=1/1) &49.10 &4.4 &18.4& 21.6 \\
   Single Domain(k=1/4) &43.37 &3.9 &11.58 &18.8\\
   Single Domain(k=1/16) &35.90 &3.2 &12.19 & 16.6\\
   Single Domain(k=1/64) &22.71 &3.2 &9.14& 15.8\\
   Single Domain(k=1/256) &12.7 &2.0 &5.48 &15.8\\

    \midrule

    \multicolumn{1}{l}{\textit{Mixed domain (Scaling)}} \\
    \midrule

   Mixed Domain(k=1/1) &47.53 &3.6 & 14.63&19.4 \\
   Mixed Domain(k=1/4) &41.98 &3.2 & 9.14& 18.8\\
   Mixed Domain(k=1/16) &32.97 &2.4 & 9.16&18.4\\
   Mixed Domain(k=1/64) &25.77 &2.4 & 14.63 &17.2 \\
   Mixed Domain(k=1/256) &14.78 &3.0 & 11.37&16.6 \\

    \midrule
    \multicolumn{1}{l}{\textit{Individual domain}} \\
    \midrule
    General only& 11.1& 2.9 &10.4 &	 1.0  \\
    Math only& 49.10 & 4.4 &6.71 & 9.0 \\
    Code only& 4.51&  1.0 & 18.40 & 21.6 \\
    \midrule
  \multicolumn{1}{l}{\textit{Different Training Strategies}} \\
  \midrule
     Multi-task learning &\textbf{47.53}& \textbf{3.6} & \underline{14.63}   & \underline{19.4}	 \\
    Sequential Training &31.39 &2.0 &15.85    &	 15.8  \\
    Mixed Sequential Training & 32.6 &2.5  &15.24  &16.6 \\
   DMT (k=1/256)  & \underline{41.92} &\textbf{3.6} & \textbf{17.68}   &\textbf{19.8}  \\
    \bottomrule
    
  \end{tabular}}
  \caption{ The detailed results of LLaMA-7B, 13B with different training strategies on OOD benchmarks.}
  \label{tab:other_benchmark}

\end{table*}

(1) In the individual domain, LLaMA shows a positive correlation between performance in MATH and MBPP and the data volume (consistent with RQ1).

(2) Comparing the individual and mixed domains, LLaMA-7B exhibits a trade-off between high-resource performance conflict and low-resource performance gain in both MATH and MBPP (consistent with RQ2).

(3) Considering the general ability results shown in Table 1, we can observe that DMT maintains competitive results in MATH and MBPP while prioritizing general abilities. This further validates the effectiveness of DMT (consistent with RQ4).

\section{Visualization of Different Layers}
\label{sec:app_g}
\begin{figure*}[h]
\centering
\begin{subfigure}[t]{0.33\linewidth}
\centering
\includegraphics[width=2in]{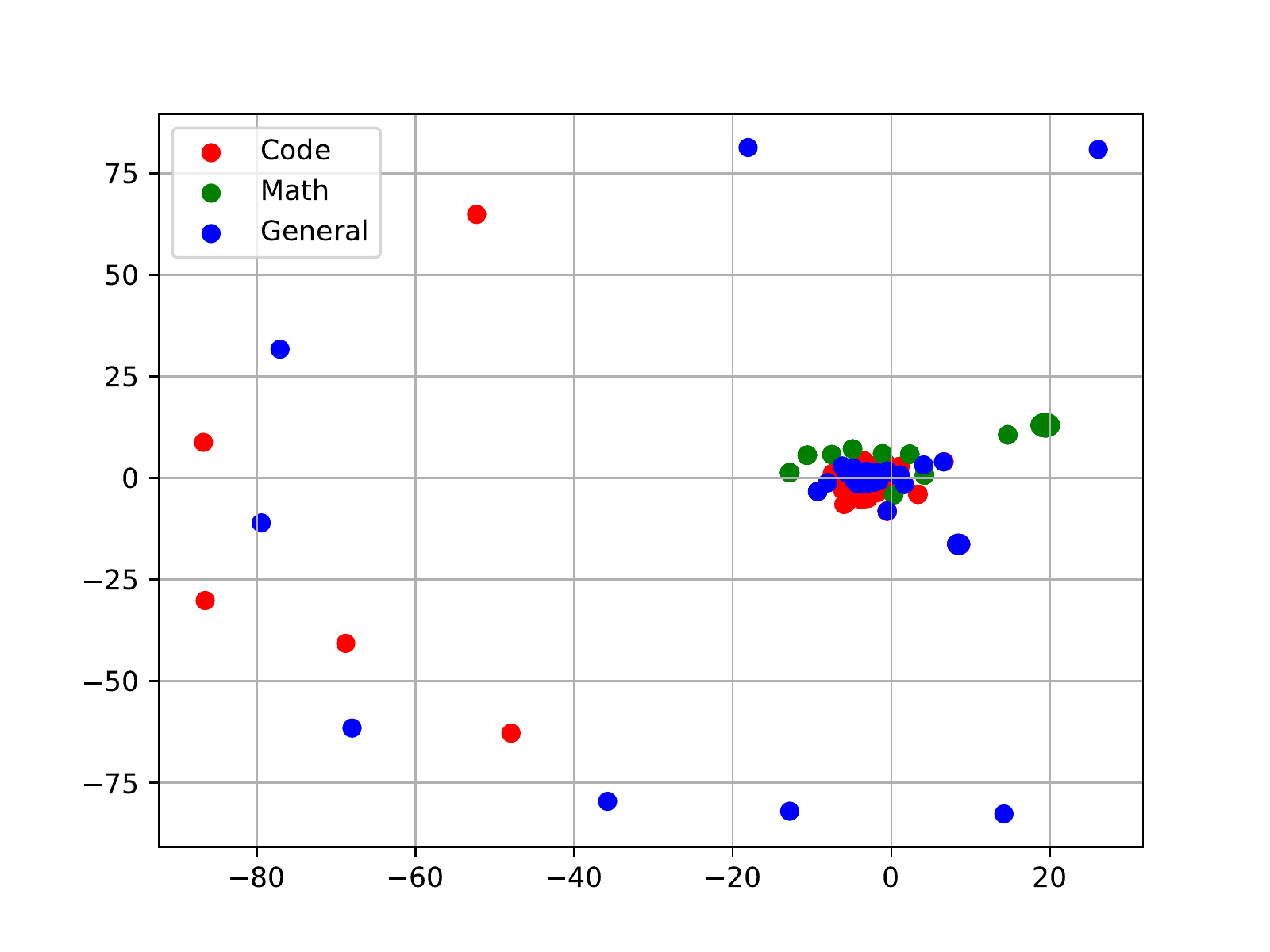}
% \caption{1}
\end{subfigure}%
\begin{subfigure}[t]{0.33\linewidth}
\centering
\includegraphics[width=2in]{figures/tsne_plot_base13b.pdf}
% \caption{2}
\end{subfigure}%
\begin{subfigure}[t]{0.33\linewidth}
\centering
\includegraphics[width=2in]{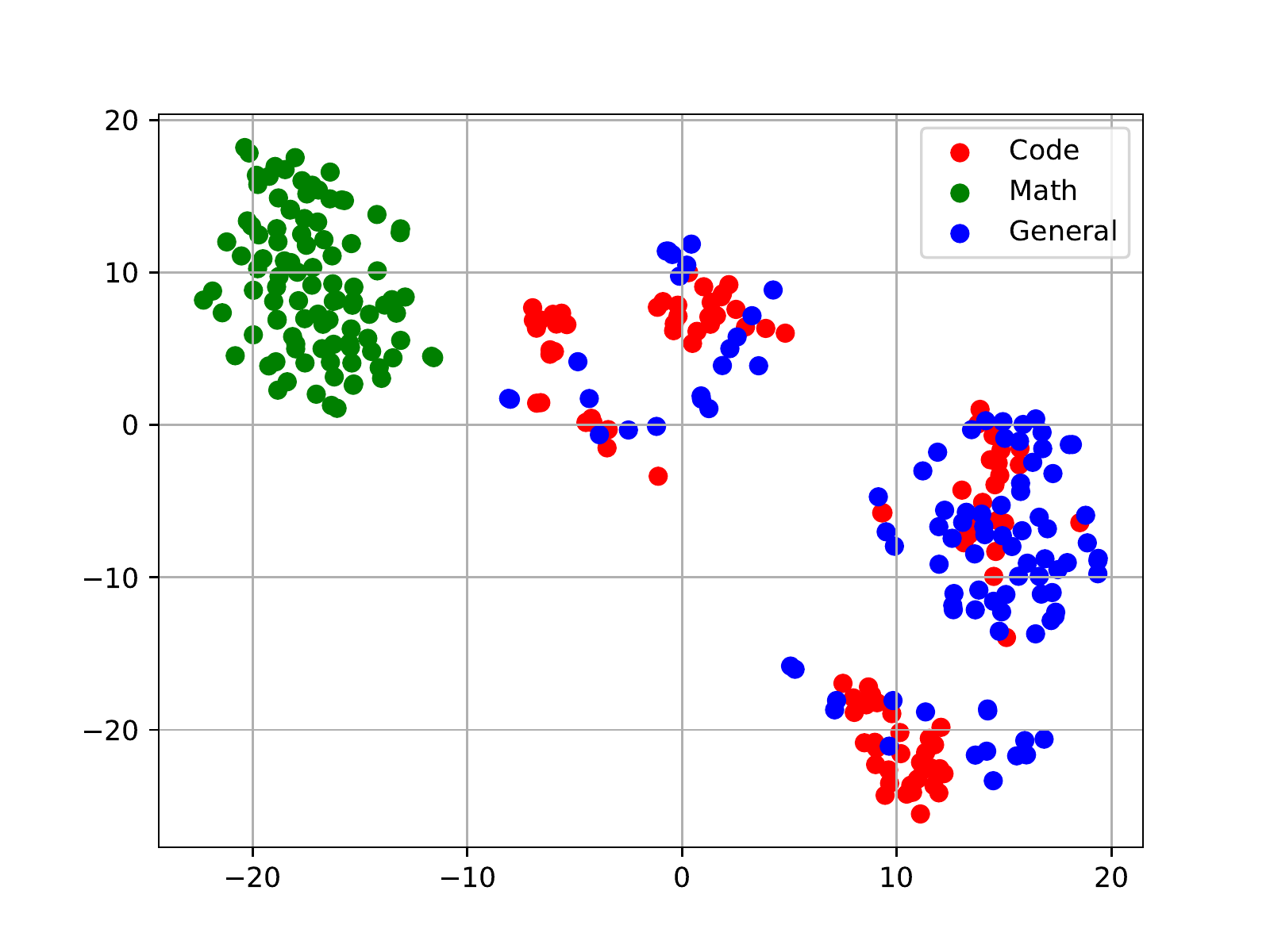}
% \caption{3}
\end{subfigure}
\caption{From left to right are the visualization results of starting layer (Layer1), middle layer (Layer15), and ending layer (Layer31) on LLaMA-7B.}
\label{fig:diff-tsne1}
\end{figure*}

\begin{figure*}[h]
\centering
\begin{subfigure}[t]{0.33\linewidth}
\centering
\includegraphics[width=2in]{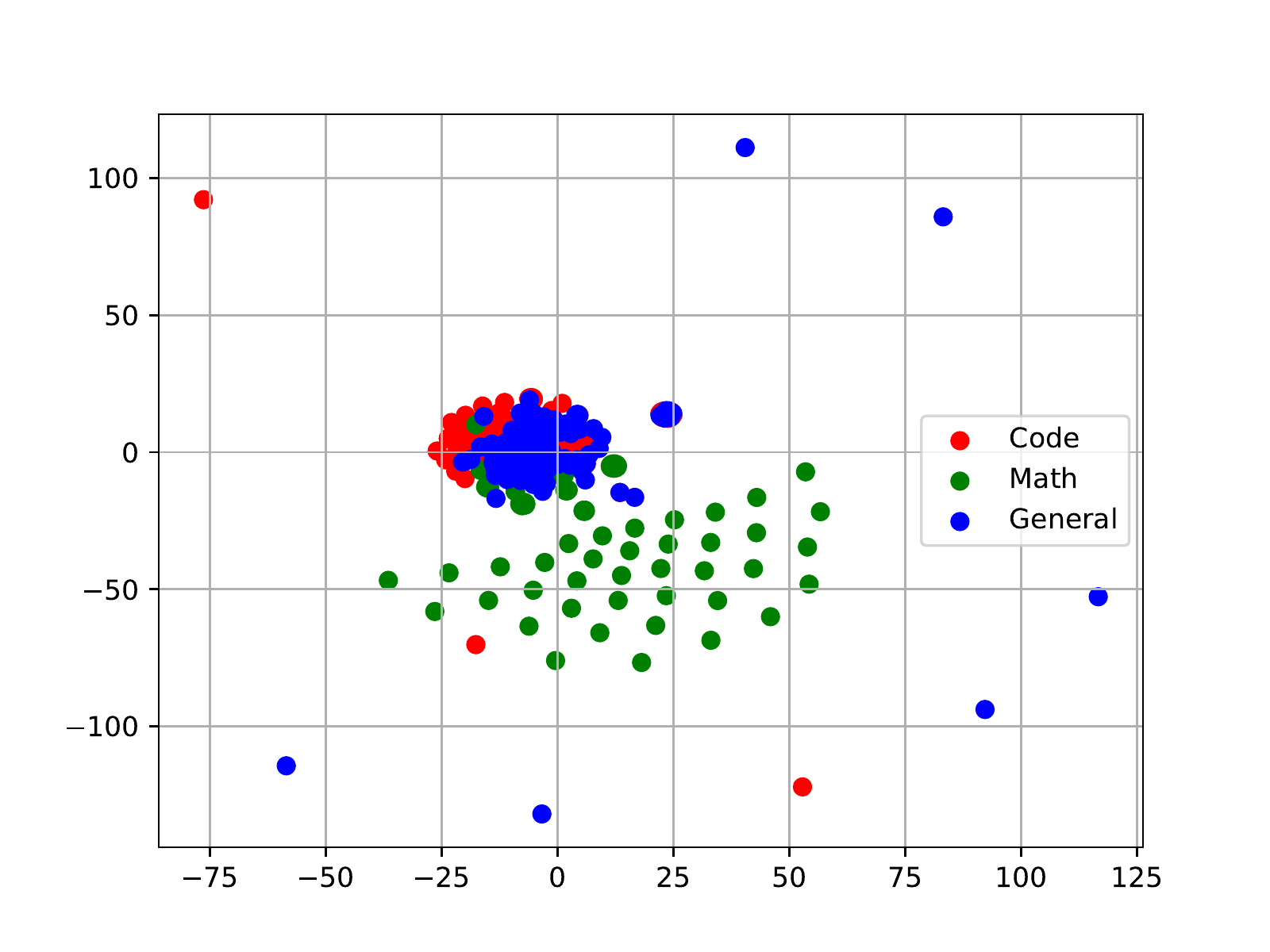}
% \caption{1}
\end{subfigure}%
\begin{subfigure}[t]{0.33\linewidth}
\centering
\includegraphics[width=2in]{figures/tsne_plot_dmt13b.pdf}
% \caption{2}
\end{subfigure}%
\begin{subfigure}[t]{0.33\linewidth}
\centering
\includegraphics[width=2in]{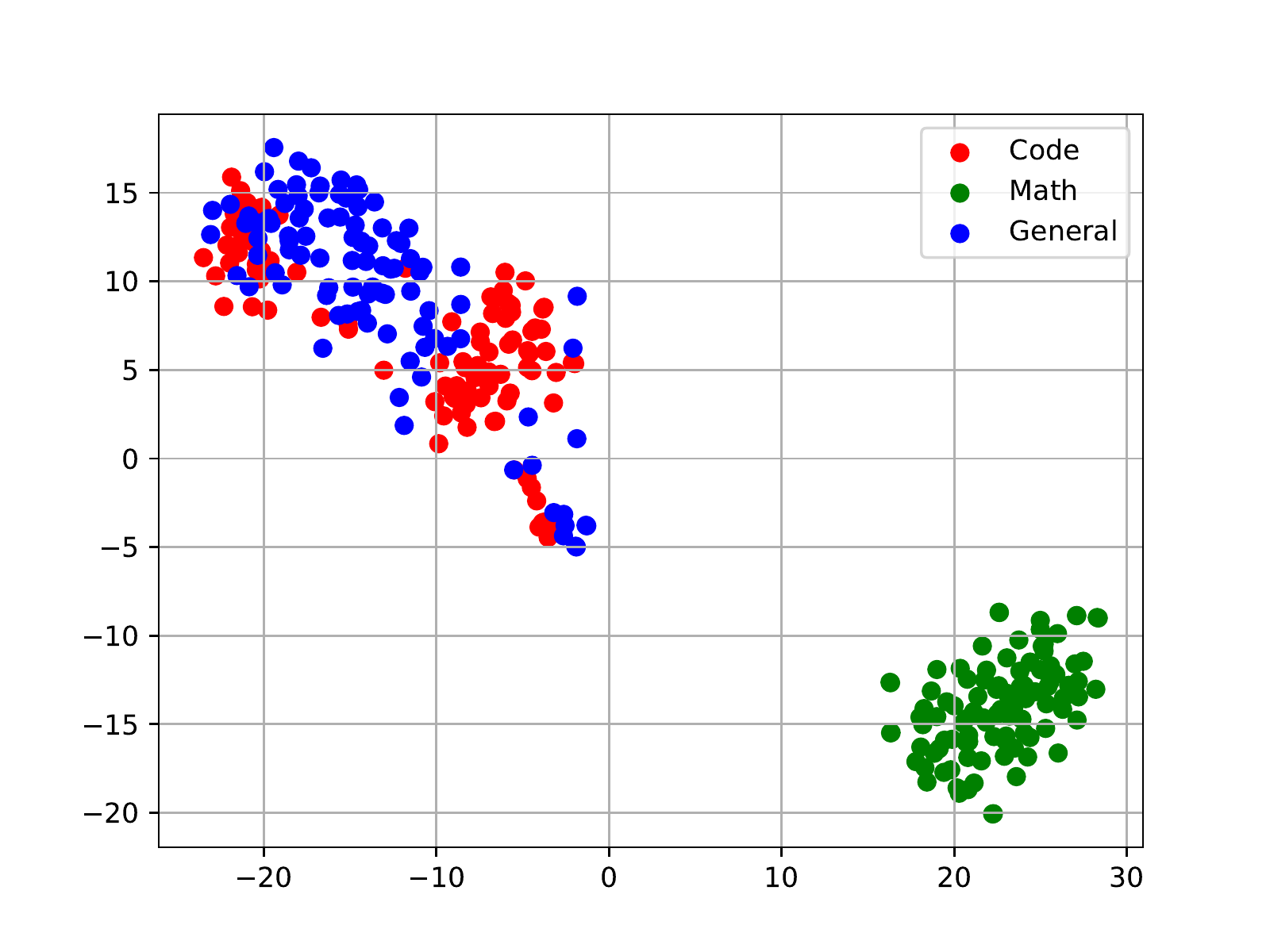}
% \caption{3}
\end{subfigure}
\caption{From left to right are the visualization results of starting layer (Layer1), middle layer (Layer15), and ending layer (Layer31) on LLaMA-7B with DMT(k=1/256) strategy.}
\label{fig:diff-tsne2}
\end{figure*}

In this section, we compared the visualization results of the baseline model of LLaMA-13B and DMT (k=1/256) in the starting layer (Layer1), middle layer (Layer15), and ending layer (Layer31) in Figure \ref{fig:diff-tsne1} and \ref{fig:diff-tsne2}.

The visualization result of the starting layer are relatively chaotic, while the visualization results of the middle layer and the ending layer are clearer. And the results of the middle layer and the last layer are consistent in pointing out that both base model and model with DMT strategy exhibit a certain level of separation in the mathematical data representations, there remains a certain degree of overlap between the representations of code and general samples.

\section{Equal Data Amount VS. Equal Data Proportion}
\label{sec:app_h}
In a realistic SFT phrase for training general LLM, the data amount for different abilities is likely to differ. Therefore, instead of controlling the same amount of data, we select to mix datasets with the same proportion of subsets to better simulate real-world scenarios in above experiments. In addition, We further supplement the experimental results using different abilities mixed with the equal data amount and compare them with the results using the equal subset proportion in Table \ref{tab:equal_data}. 

\textbf{Equal Data amount Setting:} we utilize the data amount of GSM8k RFT as the baseline. We sampled data with proportions of {1/16, 1/64, 1/256}, and mixed samples of the same data amount from Code alpaca and ShareGPT.

\textbf{Equal Proportion Setting:} we sampled data with proportions of {1/16, 1/64, 1/256} according to the subset proportions of each dataset and mixed them, which is aligned with the setup in RQ2.

It can be observed that there is not a significant difference in the results of the three benchmark tests between the two settings. Therefore, these findings do not significantly impact the main experimental conclusions presented in the paper.

\begin{table*}[h]
  \centering
  \small
  \renewcommand{\arraystretch}{1.1}
      {
  \begin{tabular}{l|ccc}
    \toprule
    % \multicolumn{3}{c}{LLaMA-33B} \\
    % \cmidrule(lr){2-4}

    Methods &GSM8K & HumanEval & MT-Bench \\
    \midrule
    % \multicolumn{1}{l}{\textit{Individual domain}} \\
    % \midrule

   Mixed Domain(k=1/16, Equal Amount) &34.49 &9.14 &5.49 \\
  Mixed Domain(k=1/64, Equal Amount) &25.02 &13.54 &5.21 \\
   Mixed Domain(k=1/256, Equal Amount) &16.7 &11.54 &4.63\\
   
    \midrule
    
   Mixed Domain(k=1/16, Equal Proportion)& 32.97 & 9.16  &5.52 \\
  Mixed Domain(k=1/64, Equal Proportion)& 25.77 &14.63   &5.24 \\
  Mixed Domain(k=1/256, Equal Proportion)& 14.78 & 11.37   &4.41 \\

    \bottomrule
    
  \end{tabular}}
  \caption{Comparative experiment between equal data amounts and equal subset proportions of different SFT abilities on LLaMA-7B}
  \label{tab:equal_data}

\end{table*}

\section{Comparison Experiment of Different Training Sequences}
\label{sec:app_i}
To investigate the impact of training order on different SFT abilities, we have conducted additional experiments with six different training orders. The results and analysis of these experiments are provided in Table \ref{tab:diff_seq}. Based on our findings, we conclude the following:

\begin{table*}[h]
  \centering
  \small
  \renewcommand{\arraystretch}{1.1}
      {
  \begin{tabular}{lccc}
    \toprule
    % \multicolumn{3}{c}{LLaMA-33B} \\
    % \cmidrule(lr){2-4}

    Methods &GSM8K & HumanEval & MT-Bench \\
    \midrule
    % \multicolumn{1}{l}{\textit{Individual domain}} \\
    % \midrule
   Code $\rightarrow$ Math $\rightarrow$ General  &31.39 &15.85 &\textbf{5.72} \\
   Math $\rightarrow$ Code $\rightarrow$ General  &29.71 &15.85 &\textbf{5.65} \\
   Code $\rightarrow$ General $\rightarrow$ Math&\textbf{48.21} &9.75 & 4.7\\
   General $\rightarrow$ Code $\rightarrow$ Math  &\textbf{48.21} &7.9 & 4.59\\
   
   General $\rightarrow$ Math $\rightarrow$ Code   &37.60 &\textbf{15.85} &3.79 \\
   Math $\rightarrow$ General $\rightarrow$ Code  &26.45 &\textbf{16.46 }&3.68 \\

    \bottomrule
    
  \end{tabular}}
  \caption{Results of different sequential training for LLaMA-7B}
  \label{tab:diff_seq}

\end{table*}

1. The SFT ability trained in the final stage tend to retain relatively good performance.

2. If general and code abilities are trained in the first two stages, there is a noticeable performance decrease in code capability, while math capability does not show significant impact. One possible reason is that the task format of code generation and general ability exhibits similar data distributions (as discussed in RQ3 and Discussion1). This can result in a more severe catastrophic forgetting phenomenon during continuous fine-tuning.

\section{Detailed Results of experiments}

\subsection{Results of Different Random Seeds}

For each dataset, we employed random selection by utilizing a random function with three distinct seeds for sampling. Subsequently, we conducted a comparative analysis of the results obtained from different subsets on the three benchmark tests. The specific details are presented in Table \ref{tab:random_seed}. It can be observed that DMT maintains its superiority under three different random seed settings. The influence of different subsets on experimental results is not a key factor and does not affect the overall trend.

\begin{table*}[t]
  \centering
  \renewcommand{\arraystretch}{1.3}
  \resizebox{\textwidth}{!}{
  \begin{tabular}{lcccccccccc}
    \toprule
    \multirow{2}{*}{Methods}&\multicolumn{3}{c}{LLaMA -7B}&\multicolumn{3}{c}{LLaMA -13B}&\multicolumn{3}{c}{LLaMA -33B} \\
    \cmidrule(lr){2-4}
    \cmidrule(lr){5-7}
    \cmidrule(lr){8-10}
    & GSM8K & HumanEval & MT-Bench& GSM8K & HumanEval & MT-Bench& GSM8K & HumanEval & MT-Bench \\
    \midrule
  
    \multicolumn{1}{l}{\textit{Different Training Strategies}} \\
     Multi-task learning &\textbf{47.53} & 14.63 &5.76 & \textbf{50.94 }  & 19.50	 & 5.73& \textbf{56.69} &18.9 &6.07   \\
    Sequential Training &31.39 &15.85&  5.72 & 39.12  &\underline{20.12}	  & 5.93 &   47.27   &\underline{24.80} & \textbf{6.73}   \\
    Mixed Sequential Training & 32.60   &15.24	  & 6.02& 40.48 & 18.30 & 5.93 & 44.24  &24.4   & 6.43 \\

    \midrule
    DMT(k=1/256,random seed=1) & 41.92& \underline{17.68} \ &  \underline{6.08} & \underline{46.47}  &19.50 &  6.03 & \underline{56.36} &25.00& \textbf{6.73}  \\

    DMT(k=1/256,random seed=2) & 41.31& \underline{17.68} \ & 6.02 & 45.85  &18.90 &  \underline{6.08} & 55.64  &\underline{24.80}   & \underline{6.71}    \\

    DMT(k=1/256,random seed=3) & \underline{42.03}& \textbf{18.21} \ & \textbf{6.13} & 46.22  &\textbf{20.52} &  \textbf{6.10} & 56.12  &\textbf{25.30}   & \textbf{6.73}    \\
    \bottomrule
  
  \end{tabular}}
  \caption{The results of LLaMA-7B, 13B, 33B under different training strategies on three benchmarks. We tested the results of DMT on randomly sampling k proportion of specified data under three random seeds.}
  \label{tab:random_seed}
  \vspace{-0.2cm}
  
\end{table*}

\subsection{Results of Single Source and Mixed Source}
In Table \ref{tab:app_single_mix1} and Table \ref{tab:app_single_mix2}, we report the detailed comparative results between mix domains and individual domains for LLaMA-7B, 3B and 33B, as the supplemental results in RQ2.

\subsection{Results of Data Ratio (k)}
In Table \ref{tab:app_dataratio}, we report The detailed results of the data ratio (k) between specific abilities and general abilities on three benchmarks, as the supplemental results in RQ3.

\subsection{Results of Specialized Data Amount of DMT}
In Table \ref{tab:app_dmt_scaling}, we report The detailed results of LLaMA-7B, 13B, 33B with different training strategies on three benchmarks, as the supplemental results in RQ4.

\subsection{Results of MT-Bench}
\begin{figure*}[h]
\centering
\begin{subfigure}[t]{0.33\linewidth}
\centering
\includegraphics[width=\linewidth]{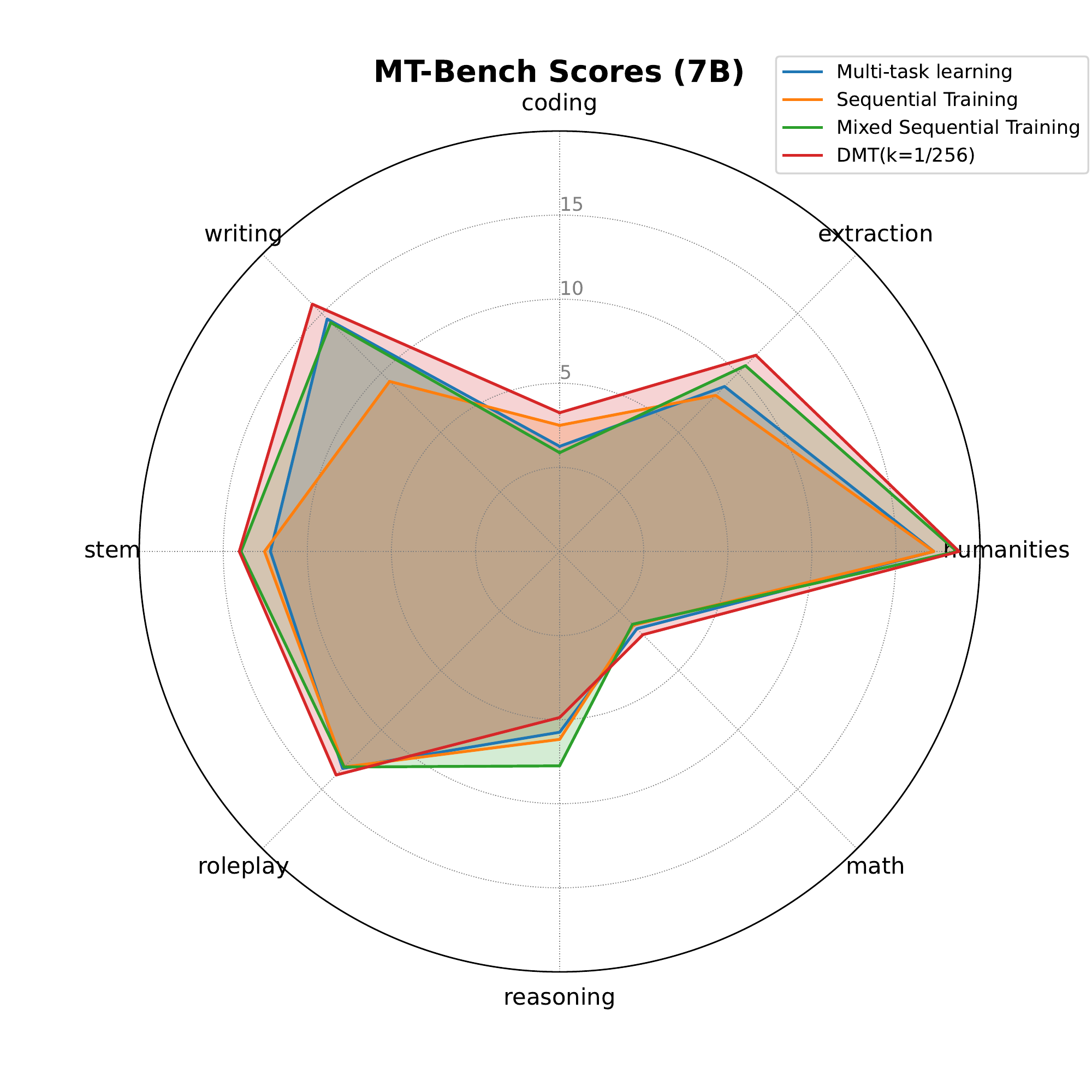}
\caption{LLaMA-7B}
\label{fig:mt7b}
\end{subfigure}%
\begin{subfigure}[t]{0.33\linewidth}
\centering
\includegraphics[width=\linewidth]{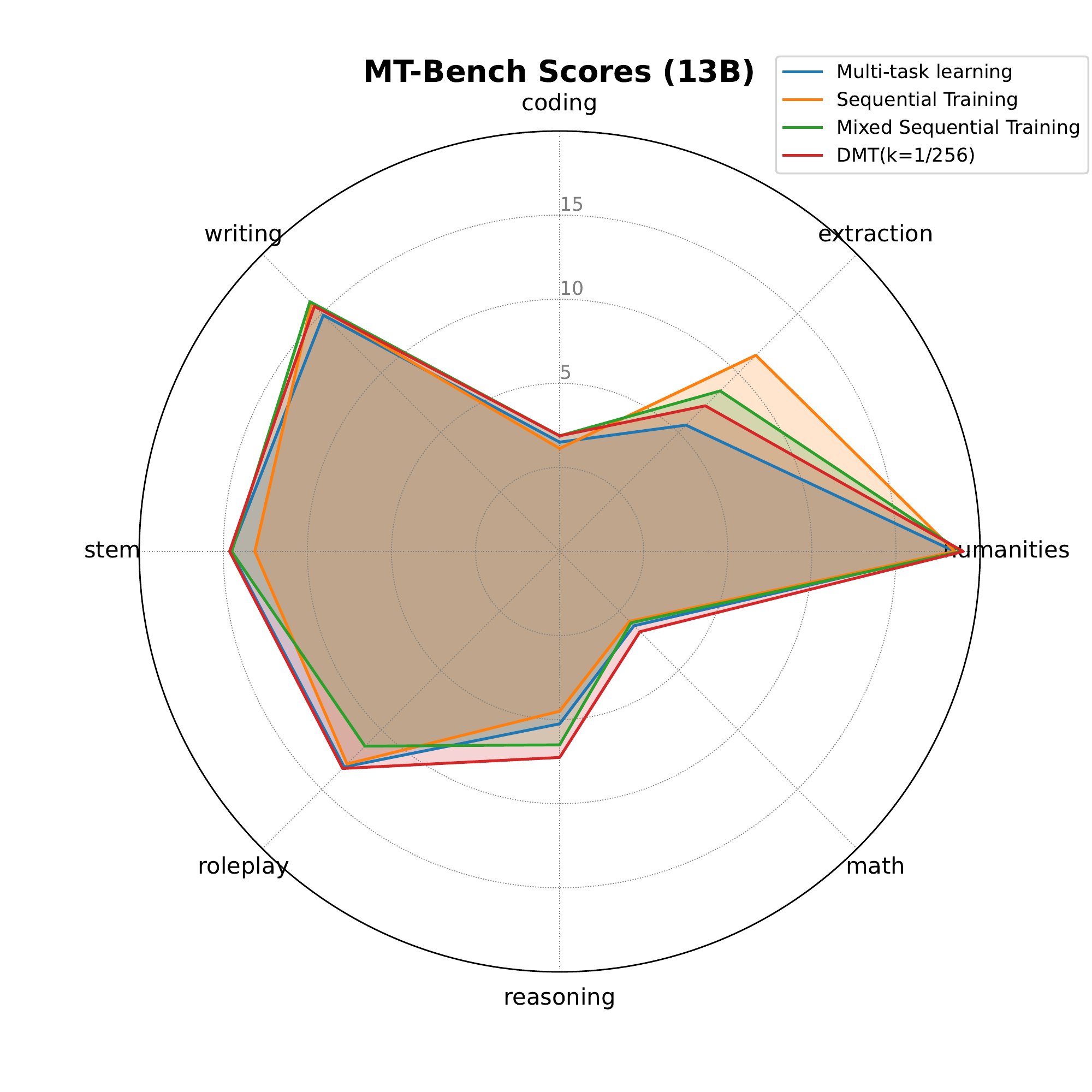}
\caption{LLaMA-13B}
\label{fig:mt13b}
\end{subfigure}%
\begin{subfigure}[t]{0.33\linewidth}
\centering
\includegraphics[width=\linewidth]{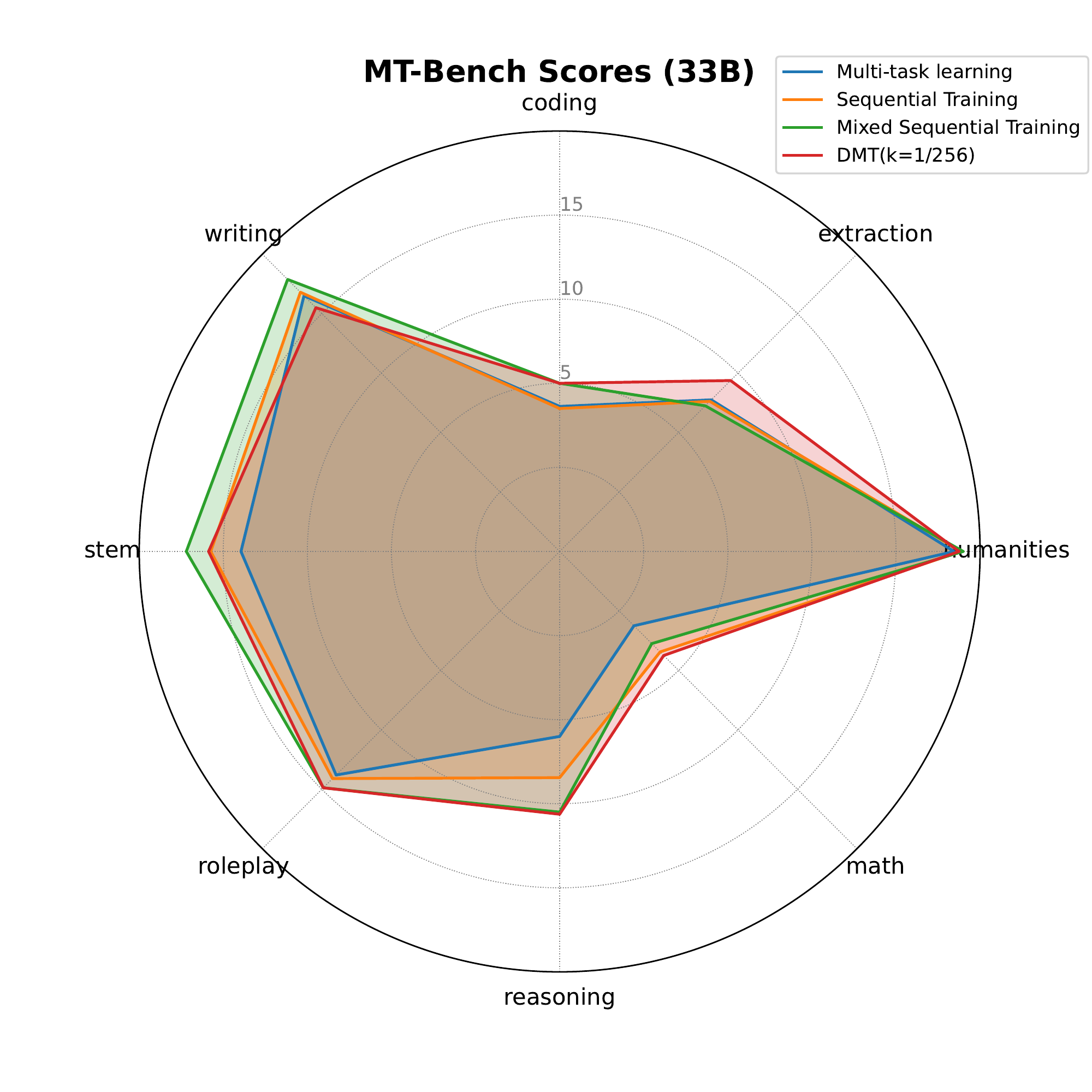}
\caption{LLaMA-33B}
\label{fig:mt33b}
\end{subfigure}
\caption{The detailed results of LLaMA-7B, 13B, 33B with different training strategies on MT-Bench.}
\label{fig:app-mtbench-detail}
\end{figure*}

In Figure \ref{fig:app-mtbench-detail}, we report detailed results of LLaMA-7B, 13B, 33B with different training strategies on MT-Bench, which include coding, extraction, humanities, math, reasoning, roleplay, stem and writing abilities.

\subsection{Supplemental Results for Discussion}
\begin{figure*}[h]
\centering
\begin{subfigure}[t]{0.5\linewidth}
\centering
\includegraphics[width=2in]{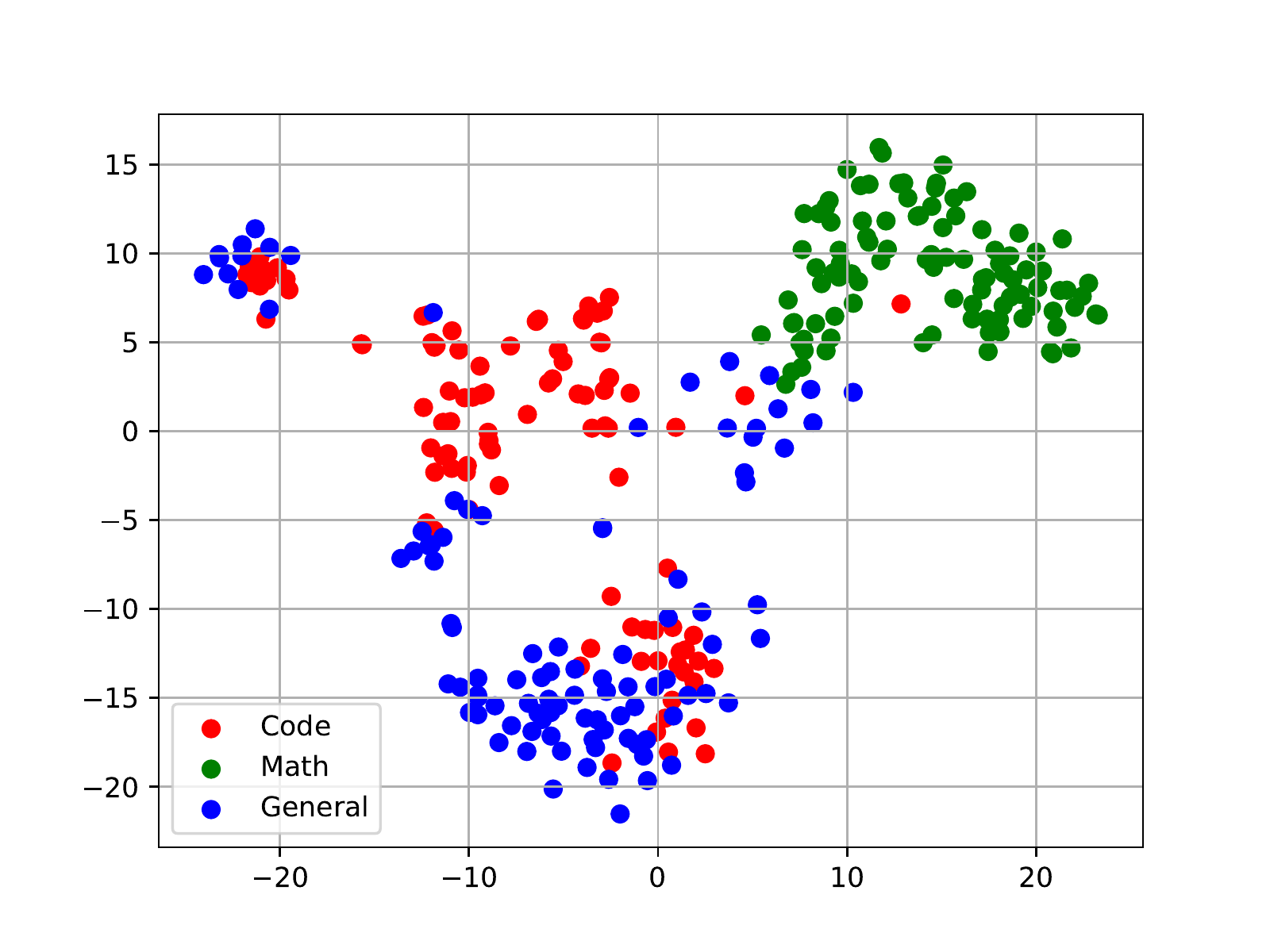}
% \caption{1}
\end{subfigure}%
\begin{subfigure}[t]{0.5\linewidth}
\centering
\includegraphics[width=2in]{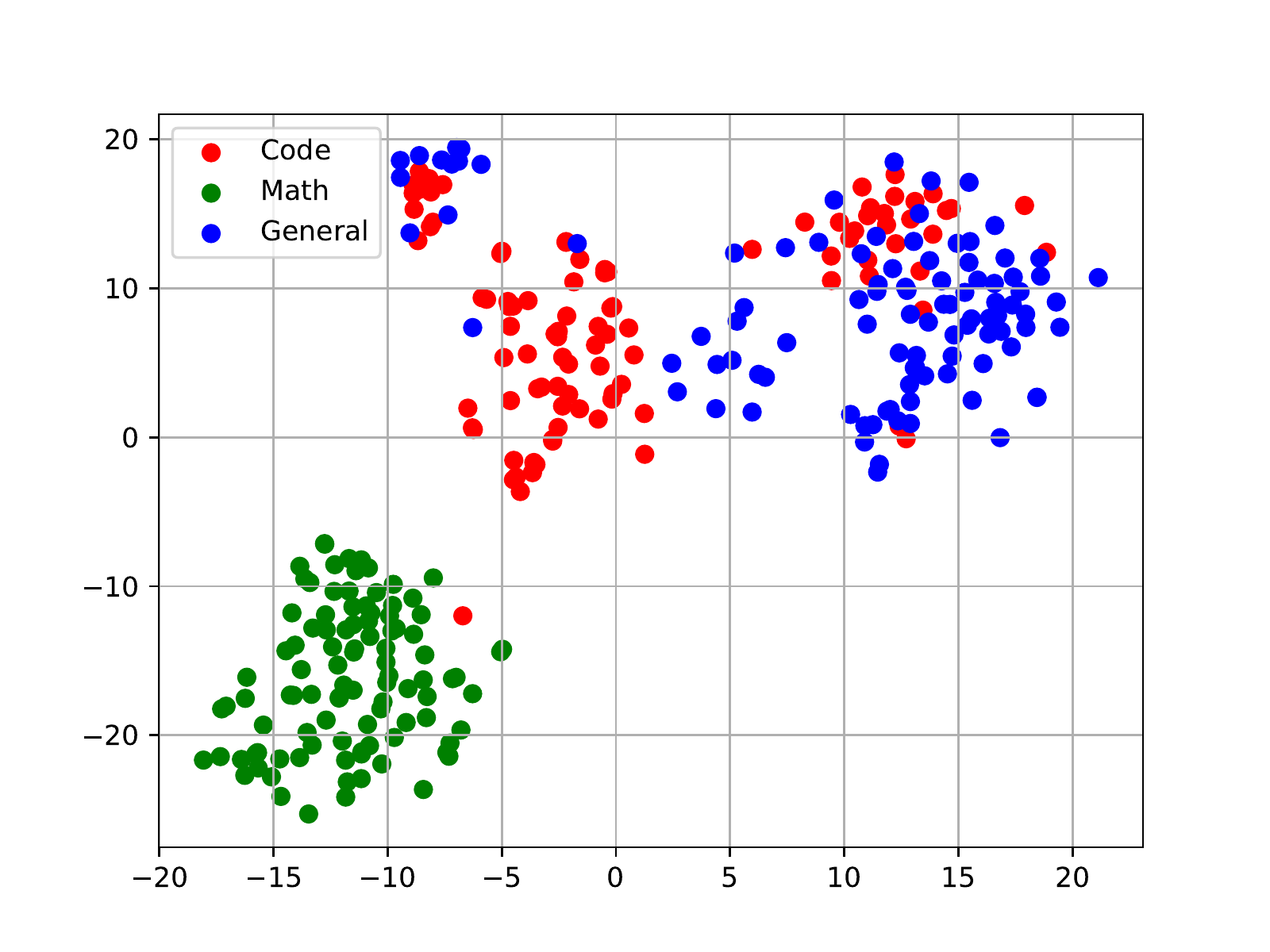}
% \caption{2}
\end{subfigure}%

% \begin{subfigure}[t]{0.33\linewidth}
% \centering
% \includegraphics[width=1.9in]{figures/dmt_scaling_7b.pdf}
% % \caption{2}
% \end{subfigure}%

\caption{Figures show the t-SNE visualizations of LLaMA-7B and LLaMA-7B with
DMT(k=1/256) stategy. }
\label{fig:discuss12}
\end{figure*}

In Figure \ref{fig:discuss12}, we report the t-SNE visualizations of LLaMA-7B and LLaMA-7B with DMT(k=1/256) strategy. What's more, the bottom figure represents the scaling relationship of LLaMA-7B with DMT(k=1/256) under different values of K.

Moreover, in Table \ref{tab:app_instag}, we report The detailed results of LLaMA-7B, 13B, 33B with different training strategies on three benchmarks, as the supplemental results in RQ4.

%---------------------------individual domain----------------------------------
\begin{table*}[h]
  \centering
  \small
  \renewcommand{\arraystretch}{1.1}
      {
  \begin{tabular}{l|cccccccccc}
    \toprule
    \multirow{2}{*}{Methods}&\multicolumn{3}{c}{LLaMA-7B}&\multicolumn{3}{c}{LLaMA-13B} \\
    \cmidrule(lr){2-4}
    \cmidrule(lr){5-7}
    & GSM8K & HumanEval & MT-Bench& GSM8K & HumanEval & MT-Bench& \\
    \midrule
    % \multicolumn{1}{l}{\textit{Individual domain}} \\
    % \midrule
   
  Single(k=1) & 49.10&  18.4 & 5.88 & 51.4 & 18.4& 6.13 \\
  Single(k=1/4) & 43.37& 11.58 &  5.85 &48.59 &13.41 & 6.03 \\
  Single(k=1/16) & 35.90&  12.19 & 5.61 & 43.00 &12.80 & 5.66\\
  Single(k=1/64) & 22.71& 9.14 & 5.11 & 27.40 &12.20 & 5.24\\
  Single(k=1/256) & 12.70& 5.48 & 4.00  & 18.40 &10.36  & 2.95 \\
  
  \hline
  Mix(k=1)&47.53 & 14.63  &5.76    &50.49 &17.10 & 5.73  \\
  Mix(k=1/4)& 41.98 & 9.14  & 5.48   & 48.52  &14.00 & 5.61 \\
  Mix(k=1/16)& 32.97 & 9.16  & 5.22   &40.63  &14.60 & 5.52 \\
  Mix(k=1/64)& 25.77 &14.63   & 5.27     &33.2  &17.68 & 5.24\\
  Mix(k=1/256)& 14.78 & 11.37   & 4.11   &24.94  &12.19 &  4.4 \\
    
    \bottomrule
    
  \end{tabular}}
  \caption{Comparative experiments between mix domains and individual domains for LLaMA-7B,
13B.}
  \label{tab:app_single_mix1}

\end{table*}
%---------------------------individual domain---------------------------------

%---------------------------individual domain----------------------------------
\begin{table*}[h]
  \centering
  \small
  \renewcommand{\arraystretch}{1.1}
      {
  \begin{tabular}{l|ccc}
    \toprule
    % \multicolumn{3}{c}{LLaMA-33B} \\
    % \cmidrule(lr){2-4}

    Methods &GSM8K & HumanEval & MT-Bench \\
    \midrule
    % \multicolumn{1}{l}{\textit{Individual domain}} \\
    % \midrule
   
    Single(k=1) & 57.91& 26.82&6.63\\
Single(k=1/4) &56.10 &25.61&6.66 \\
  Single(k=1/16) &54.60 &21.95&6.17\\
  Single(k=1/64)  &44.60 & 18.59 &5.99\\
  Single(k=1/256)&29.21 & 14.02 & 2.3 \\
  
  \hline
  Mix(k=1) &56.69 & 18.9 & 6.07 \\
  Mix(k=1/4) & 54.54 &22.56&  5.92\\
 Mix(k=1/16)&53.33 & 26.82& 6.26\\
  Mix(k=1/64)&46.66  & 18.6&5.73\\
  Mix(k=1/256) & 36.54 & 17.68 & 4.58\\
    
    \bottomrule
    
  \end{tabular}}
  \caption{Comparative experiments between mix domains and individual domains for LLaMA-33B.}
  \label{tab:app_single_mix2}

\end{table*}
%---------------------------individual domain---------------------------------

%------------------------------7B----------------------------------

\begin{table*}[h]
       \centering
    \small
    \renewcommand{\arraystretch}{1.5}
    \begin{tabular}{l|cccc}
  \hline
  Model size & GSM8K & HumanEval & MT-Bench \\

 %  \hline
 % 1 Mix(code,math,general)&47.53 & 14.63  &5.76       \\
 % 1/4 Mix(code,math,general)& 41.98 & 9.14  & 5.48    \\
 %  1/16 Mix(code,math,general)& 32.97 & 9.16  & 5.22     \\
 % 1/64 Mix(code,math,general)& 25.77 &14.63   & 5.27    \\
 % 1/256 Mix(code,math,general)& 14.78 & 11.37   & 4.11 \\

%[11.37, 14.63, 9.16, 9.14, 14.63]
 \hline
  Mix[(code,math),1 general]&47.53 & 14.63 	 & 5.76  \\
  Mix[(code,math),1/4 general]&48.44   &15.85	 & 5.73 \\
  Mix[(code,math),1/16 general]&47.99   &15.24	 &  5.27 \\
Mix[(code,math),1/64 general]&47.23   &14.63	 &  5.16 \\
 Mix[(code,math),1/256 general]&48.52   &16.46	 & 4.69  \\

\hline
 Mix[1(code,math),general]&47.53 & 14.63 	 &5.76    \\
 Mix[1/4(code,math),general]&41.31   &10.97	 &  5.81  \\
 Mix[1/16(code,math),general]&33.20   &11.58	 & 5.76 \\
 Mix[1/64(code,math),general]&25.17   &12.19	 &  5.84 \\
 Mix[1/256(code,math),general]&16.52   &9.14	 &   5.82  \\

\hline
  Mix[1(code,math),1/64general] &47.68 & 14.63 & 5.09 \\
  Mix[1/4(code,math),1/64general] & 43.29 & 12.19  & 5.07 \\
  Mix[1/16(code,math),1/64general] &33.81 & 12.19 &5.17 \\
  Mix[1/64(code,math),1/64general] & 26.23 & 12.19& 5.12 \\
  Mix[1/256(code,math),1/64general] & 18.27 &10.36  & 5.12 \\

\hline
\end{tabular}
\caption{The detailed results of the data ratio (k) between specific abilities and general abilities on three benchmarks.}
\label{tab:app_dataratio}
\end{table*}

%------------------7B--------------------------

%---------------------------DMT detail of 7b 13b----------------------------------
\begin{table*}[h]
   \small
 \centering
  \renewcommand{\arraystretch}{1}
  \resizebox{0.9\textwidth}{!}{
  \begin{tabular}{lccccccc}
    \toprule
    \multirow{2}{*}{Methods}&\multicolumn{3}{c}{LLaMA-7B}&\multicolumn{3}{c}{LLaMA-13B} \\
    \cmidrule(lr){2-4}
    \cmidrule(lr){5-7}
   
    & GSM8K & HumanEval & MT-Bench& GSM8K & HumanEval & MT-Bench\\
    \midrule
    \multicolumn{1}{l}{\textit{Individual domain}} \\
    % \midrule
   General only& 11.10&	10.42& 5.88 &14.02 &16.40 &	 6.13    \\
    Math only& 49.10 & 6.71 & 2.53& 51.40 & 12.8 & 2.54 \\
    Code only& 4.51& 18.40 & 4.30& 5.15& 17.1 & 3.53  \\

    \midrule
  
    \multicolumn{1}{l}{\textit{Different Training Strategies}} \\
     Multi-task learning &47.53 & 14.63 &5.76 & 50.94  & 19.5	 & 5.73 \\
    Sequential Training &31.39 &15.85&  5.72 & 39.12  &20.12	  & 5.93  \\
    Mixed Sequential Training & 32.6   &15.24	  & 6.02& 40.48 & 18.30 & 5.93 \\
    \midrule
  DMT (k=1) & 45.79&  14.02 & 5.63 &50.49  &16.46 & 5.76\\
  DMT (k=1/4) & 48.37 & 13.41 &  5.69& 50.18  &18.9 & 5.83 \\
  DMT (k=1/16)  & 43.3 &  15.24 &5.78 &48.59  &18.9 & 5.96 \\
  DMT (k=1/64)  & 42.53& 15.85 & 6.01&47.61  &15.24 & 6.03   \\
  DMT (k=1/256)  & 41.92& 17.68  & 6.08 &46.47  &19.5 &  6.03 \\
    \bottomrule
    
  \end{tabular}}
  \caption{ The detailed results of LLaMA-7B, 13B with different training strategies on three benchmarks.}
  \label{tab:app_dmt_scaling}

\end{table*}

% %------------------DMT detail 7b 13b --------------------------

% %-----------mtbench all------------
% \begin{figure}[h]
%     \centering
%     \small    \includegraphics[width=\linewidth]{figures/mt7b.pdf}
%     \caption{The detailed results of LLaMA-7B on MT-Bench}
%     \label{fig:mt1}
% \end{figure}
% %-----------mtbench alls------------

% %-----------mtbench all------------
% \begin{figure}[h]
%     \centering
%     \small    \includegraphics[width=\linewidth]{figures/mt13b.pdf}
%     \caption{The detailed results of LLaMA-13B on MT-Bench}
%     \label{fig:mt2}
% \end{figure}
% %-----------mtbench alls------------

% %-----------mtbench all------------
% \begin{figure}[h]
%     \centering
%     \small    \includegraphics[width=\linewidth]{figures/mt33b.pdf}
%     \caption{The detailed results of LLaMA-33B on MT-Bench}
%     \label{fig:mt3}
% \end{figure}
% %-----------mtbench alls------------

%------------------------------7B----------------------------------

\begin{table*}[h]
    \centering
    \small

    \renewcommand{\arraystretch}{1.5}
    \begin{tabular}{l|ccc}
  \hline
  Model size & GSM8K & HumanEval & MT-Bench\\
  \hline

 % \hline
  1/1 Mix(code,math,general(w/o code math)) & 49.05 &17.68 &5.80 \\
  1/4 Mix(code,math,general(w/o code math)) & 43.13 & 15.85 & 5.71 \\
  1/16 Mix(code,math,general(w/o code math)) & 36.23 &10.36 & 5.38 \\
  1/64 Mix(code,math,general(w/o code math)) & 25.62 & 10.97 & 5.21\\
 1/256 Mix(code,math,general(w/o code math)) & 15.31& 11.37 & 4.38 \\

\hline
\end{tabular}
\caption{The scaling curve after ablating code and math-related samples from ShareGPT}
\label{tab:app_instag}
\end{table*}

%------------------7B--------------------------

%----------------mt bench---------------------

\end{document}